\documentclass[times,10pt,twocolumn]{article}
\usepackage{wacv}
\usepackage{times}
\usepackage{epsfig}
\usepackage{graphicx}
\usepackage{amsmath}
\usepackage{amssymb}
\usepackage{booktabs}
\usepackage{multirow}
\usepackage[nolist]{acronym}
\usepackage{enumitem}
\usepackage{balance}
\usepackage[table]{xcolor}
\usepackage{makecell}
\usepackage{subcaption}

\wacvfinalcopy % *** Uncomment this line for the final submission

\ifwacvfinal
\usepackage[pagebackref=false,breaklinks=true,colorlinks,bookmarks=false]{hyperref}
\else
\usepackage[pagebackref=false,breaklinks=true,colorlinks,bookmarks=false]{hyperref}
\fi

\ifwacvfinal
\else
\fi

\newcommand{\trtitle}{Spatio-Temporal Analysis of Facial Actions using \\Lifecycle-Aware Capsule Networks}

\begin{document}

%%%%%%%%% TITLE
\title{\trtitle}
\author{Nikhil~Churamani$^{*}$, Sinan~Kalkan$^{\dagger}$ and Hatice~Gunes$^{*}$\\
    $^{*}$Department of Computer Science and Technology, University of Cambridge, United Kingdom\\
    $^{\dagger}$Department of Computer Engineering, Middle East Technical University, Ankara, Turkey\\
     {\tt\small\{nikhil.churamani, hatice.gunes\}@cl.cam.ac.uk, skalkan@metu.edu.tr}
}

\maketitle
%\thispagestyle{empty}
% \setlength{\belowcaptionskip}{-12pt}

%%%%%%%%% ABSTRACT
\begin{abstract}
  Most state-of-the-art approaches for Facial \acf{AU} detection rely upon evaluating facial expressions from static frames, encoding a snapshot of heightened facial activity. In real-world interactions, however, facial expressions are usually more subtle and evolve in a temporal manner requiring \ac{AU} detection models to learn spatial as well as temporal information. In this paper, we focus on both spatial and spatio-temporal features encoding the temporal evolution of facial \ac{AU} activation. For this purpose, we propose the \acf{AULA-Caps} that performs \ac{AU} detection using both frame and sequence-level features. While at the frame-level the capsule layers of \ac{AULA-Caps} learn spatial feature primitives to determine \ac{AU} activations, at the sequence-level, it learns temporal dependencies between contiguous frames by focusing on relevant spatio-temporal segments in the sequence. The learnt feature capsules are routed together such that the model learns to selectively focus more on spatial or spatio-temporal information depending upon the \ac{AU} lifecycle. The proposed model is evaluated on the commonly used BP4D and GFT benchmark datasets obtaining state-of-the-art results on both the datasets.

\end{abstract}
%---------------------------------------------------------------
\section{Introduction}

Analysing facial expressions can be highly subjective and influenced by different contextual and cultural variations~\cite{Jack7241}. To establish constants across varying cultural contexts and achieve objective evaluations for facial expressions, Ekman \etal\cite{ekman1978facial,Ekman2002FACS} developed the \ac{FACS}. Facial actions, that is, the contraction and relaxation of facial muscles are encoded as `\textit{activated}' facial \acfp{AU} that can be used to describe different facial expressions. As \ac{FACS} only encodes the activation of facial muscles, no subjective or context-sensitive affective understanding is needed. Co-activation of different \acsp{AU} reveals local relationships and hierarchies where multiple facial muscles combine to form an expression, for example, raised eyebrows (involving \acsp{AU} 1, 2) and jaw-drop (\acs{AU} 26) together signify \textit{surprise}~\cite{ekman1978facial} or communicate an intention (for example, \acs{AU} 46 for winking).
% \ac{FACS} annotates different combinations of facial muscle movements accompanying an individual's affective responses. 

Furthermore, facial muscle activation follows a temporal evolution~\cite{Pantic2005Detecting, Sariyanidi2015Automatic}, referred to in this paper as the \textit{\ac{AU} Lifecycle}. Starting from a relaxed and \textit{neutral} resting state, facial muscles start to contract, forming the \textit{onset} of an expression with complete contraction achieved in the \textit{apex} state to express peak intensity. This is followed by the relaxation of the muscles forming the \textit{offset} state before returning back to \textit{neutral}. This process may also be repeated several times for certain expressions, for example, spontaneous smiles typically have multiple \textit{apices} with a much slower \textit{onset} phase~\cite{COHN2004Timing}. Understanding this evolution is essential to understand how humans express affect and is particularly crucial for distinguishing \textit{posed} or \textit{acted} expressions from \textit{spontaneous} expressions~\cite{Valstar2007DPS}. 

Computational models for \ac{AU} detection, traditionally, have explored local spatial relationships and hierarchies between different face regions using shape-based representations or using spectral or histogram-based methods~\cite{Sariyanidi2015Automatic, zeng2009survey}. With deep learning gaining popularity, recent approaches~\cite{Chu2017Learning, Gudi2015Deep,Li2018EACNet, Shao2019Facial, shao2020spatiotemporal, Zhao2016Deep, Zhao2015Joint} have applied convolution or graph-based models to focus on learning such facial features directly from data, outperforming most traditional approaches. More recently, capsule-based computations proposed by Sabour \etal~\cite{CAPS2017} have further improved the learning of spatial dependencies in the form of facial feature \textit{primitives}. These feature primitives are sensitive to local variations and capture local dependencies between different facial regions and have been successfully applied for \ac{AU} detection and expression recognition tasks~\cite{Ertugrul2018FACSCaps, Quang2019CapsNet, rashid2018facial}.

Most approaches, however, only focus on frame-based evaluation of facial actions, relying on analysing peak-intensity frames~\cite{nott44740,Sariyanidi2015Automatic}. As a result, even though these approaches are able to detect strong \ac{AU} activations in \textit{posed} settings or when an expression is highly \textit{accentuated}, they suffer when detecting more \textit{subtle} expressions in \textit{spontaneous} and naturalistic settings~\cite{Valstar2007DPS, Yang2019FACS3D}, challenging their real-world applicability. A prevailing requirement for \textit{automatic} \acs{AU} detection is to be sensitive to the said \textit{\ac{AU} lifecycle} and include temporal information, such as motion features or correlations amongst proximal frames, along with spatial features~\cite{li2019semantic,shao2020spatiotemporal,Yang2019FACS3D}. While spatial processing is important to determine relationships between different facial regions~\cite{li2019semantic}, understanding temporal correlations between their activation patterns in contiguous frames provides essential information about the \textit{\ac{AU} lifecycle} and can be particularly useful in detecting subtle activations~\cite{Chu2017Learning,shao2020spatiotemporal,Yang2019FACS3D}. 

% In this work, we focus on the spatio-temporal analysis of facial actions, combining both spatial and spatio-temporal features to focus on the \ac{AU} lifecycle. 
% In this work, we focus on the spatio-temporal analysis of facial actions. 
Leveraging the ability of capsule networks to learn relationships between local spatial and temporal features, in this work, we propose the \acf{AULA-Caps} for multi-label \ac{AU} detection (see Figure~\ref{fig:aula_model}). \acs{AULA-Caps} is a multi-stream, lifecycle-aware capsule network trained in an end-to-end manner that not only learns spatial activation patterns within a frame but also their dynamics across several contiguous frames. To the best of our knowledge, this is the first work combining multiple capsule-based processing streams that learn spatial and spatio-temporal facial features at frame and sequence-level, simultaneously. 
We perform benchmark evaluations on the GFT~\cite{GFTDatabase} and BP4D~\cite{ZHANG2014BP4D} datasets comparing \ac{AULA-Caps} performance with the state-of-the-art methodologies on \ac{AU} detection. \ac{AULA-Caps} achieves the best F1-score for \acsp{AU} $2, 7, 17$ and $23$ with the second-best overall (average) F1-score for the GFT dataset while achieving the best F1-scores for \acsp{AU} $1, 6$ and $17$ along with the best overall F1-score for the BP4D dataset.

%The spatial processing stream learns spatial dependencies in a face image and is sensitive to co-activation patterns of different face regions while the spatio-temporal processing stream focuses on learning dependencies between contiguous frames. % The extracted feature primitives, that is, the primary capsules from each stream encoding respective feature-primitives are concatenated together and used to predict the \ac{AU} in input sequence making use of the \textit{routing-by-agreement} mechanism~\cite{CAPS2017}. We train and evaluate \ac{AULA-Caps} on popular \textit{posed} and \textit{spontaneous} \ac{AU} detection datasets. 

%work can be summed up as follows:
%\begin{enumerate}[leftmargin=0.4cm,label=\roman*)]

%    \item %We propose the \acs{AULA-Caps}, 
    % An approach close to ours is the FACS3D-Net~\cite{Yang2019FACS3D} which uses 2D and 3D \ac{CNN}-based processing but evaluates spatio-temporal features over the entire length of video-sequences. Our approach, in contrast, uses short temporal windows of contiguous frames.
%    \vspace{-2mm}
%    \item 

    % \item Evaluations on benchmark \ac{AU} datasets such as BP4D and GFT highlights the model's adaptability to different contextual settings, trained in an end-to-end manner on each of the datasets obtaining state-of-the-art results.
% \end{enumerate}
%-------------------------------------------------------------------------
%-------------------------------------------------------------------------

\section{Related Work}
\subsection{Spatial Analysis for \ac{AU} Prediction}
%-------------------------------------------------------------------------
% They compute local relationships between different parts of the face and use these to predict whenever one or more \acp{AU} are activated. 
\ac{AU} detection approaches rely on capturing spatial relationships between different face regions~\cite{Pantic2000Automatic,Pantic2004Facial,Sariyanidi2015Automatic}. Popular methods include using geometric features that track facial landmarks~\cite{IntraFace2015}, histogram-based approaches that cluster local features into uniform regions for processing~\cite{Sariyanidi2015Automatic} or using features that describe local neighbourhoods~\cite{bargal2012classification}. With the popularity of deep learning, \acs{CNN}~\cite{Ertugrul2019Cross,Gudi2015Deep,Li2018EACNet} and graph-based~\cite{li2019semantic,shao2020spatiotemporal} approaches have achieved state-of-the-art results for \ac{AU} detection due to their ability to hierarchically learn spatial features. Capsule-based computations~\cite{CAPS2017} offer an improvement as along with learning to detect different facial features, they also learn how these are arranged with respect to each other. Recent work~\cite{Ertugrul2018FACSCaps,rashid2018facial} has explored capsule-based computations for \ac{AU} detection by learning facial features that capture variations with respect to pose and/or orientation. Yet, relying only on spatial features ignores how \ac{AU} activations evolve over time, impacting performance on automatic \ac{AU}~detection~\cite{Yang2019FACS3D}.

\subsection{Spatio-Temporal Analysis for \ac{AU} Prediction}
%-------------------------------------------------------------------------

% Facial \ac{AU} activity follows a temporal activation pattern where facial muscles contract and relax over time resulting in the respective \ac{AU} activation evolving dynamically~\cite{Valstar2007DPS,Sariyanidi2015Automatic}.
% The temporal evolution of \ac{AU} activations is separated into three segments namely, \textit{onset}, \textit{apex} and \textit{offset}, each corresponding to certain intensity of \ac{AU} activation~\cite{Pantic2005Detecting}. Experienced human \ac{AU} coders analyse contiguous frames to determine \ac{AU} activations, evaluating their occurrence as well as intensity to capture even subtle changes~\cite{Yang2019FACS3D}. 

% Focusing only on spatial relationships using peak-intensity apex frames does not capture the dynamics of \ac{AU} activation and may result in poor performance of automatic \ac{AU} detection algorithms particularly in real-world settings. 
Learning dynamic spatio-temporal features provides information about the dynamics of \ac{AU} activations. A straightforward way for computing spatio-temporal features is to extract spatial features from each frame separately and use recurrent models such as the \acs{LSTM}~\cite{Hochreiter1997} to learn how these evolve with time~\cite{Chu2017Learning}. Alternatively, models may compute temporal features, such as optical flow, first and then process them using \acs{CNN}-based networks~\cite{Allaert2019OpticalFlow}. Yet, most of these approaches focus on learning spatial and temporal information sequentially. Yang \etal~\cite{Yang2019FACS3D} propose an alternative by concurrently learning spatial and temporal features, inspired by human \ac{AU} coders. However, their approach focuses on extracting spatio-temporal features from complete video sequences at once, dropping certain adjacent frames to ensure all video sequences are of the same length. Other more recent approaches learn semantic relationships between different face regions and represent these using structured knowledge-graphs, learning coupling patterns between regions using graph-based computations~\cite{li2019semantic,shao2020spatiotemporal}. 
% Such spatio-temporal learning can also be combined with spatial processing of facial frames forming a multi-stream approach aiding learning in the models. 

\begin{figure*}
    \centering
    \includegraphics[width=\textwidth]{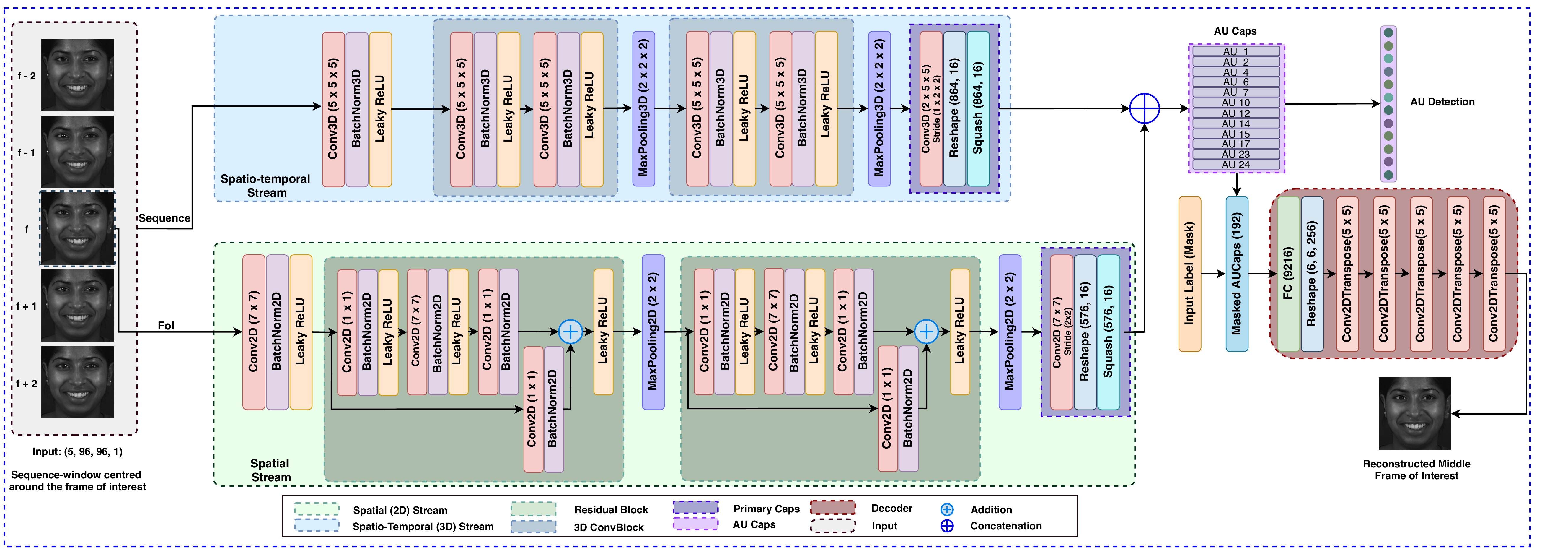}
    \caption{\acf{AULA-Caps} for Multi-label Facial Action Unit Detection.}
    \label{fig:aula_model}
    \vspace{-4mm}
\end{figure*}

\subsection{Capsule Networks}

Sabour \etal~\cite{CAPS2017} proposed the Capsule Networks that learn spatial dependencies in the form of feature \textit{primitives} by extracting features corresponding to the different regions of an input image and learning how they combine together to contribute towards solving a particular task. This ability to learn local features and their inter-dependencies makes them a good fit for \ac{AU} detection. Ertugrul \etal~\cite{Ertugrul2018FACSCaps} propose the FACSCaps model that employs capsule networks to learn pose-independent spatial feature representations from multi-view facial images for \ac{AU} detection. Rashid \etal~\cite{rashid2018facial} use capsule networks consisting of multiple convolutional operations to extract relevant spatial features from static frames before \textit{routing} them together to obtain fully-connected class capsules. A similar approach is also employed by Quang \etal~\cite{Quang2019CapsNet} applying capsule networks for micro expression recognition. These approaches, however, focus only on learning spatial features from static images.

Capsule networks have also been applied for video-based action recognition~\cite{NIPS2018_7988} that use $3$D capsules for segmenting and tracking objects across frames. However, they explore temporal relations between frames only for segmentation and ignore how these might contribute towards sequence-based predictions. Jayasekara \etal~\cite{Jayasekara2019TimeCapsCT}, on the other hand, apply capsule-based learning for time-series predictions learning to classify $1$D ECG signals by focusing on temporal dependencies. 
% \textit{temporal} features and evaluation of specific temporal patterns for classifying sequences.

In this paper, we propose a multi-stream approach that applies capsule-based computations both at frame and sequence-level, concurrently learning spatial as well as spatio-temporal dependencies from sequences of contiguous frames.

%-------------------------------------------------------------------------
%-------------------------------------------------------------------------

\section{\acf{AULA-Caps}}
\label{sec:method}

% combines spatial features extracted from a \acf{FoI} and spatio-temporal features extracted from a window of contiguous frames around the \ac{FoI} for analysing facial actions. The model
We propose the \ac{AULA-Caps} network (see Figure.~\ref{fig:aula_model}) that processes face-image sequences using two separate streams for computing spatial ($2$D) and spatio-temporal ($3$D) features. While spatial processing of a \ac{FoI}, in this case, the middle frame from each input sequence, focuses on local spatial dependencies, spatio-temporal processing investigates contiguous frames to extract features that capture dynamics of \ac{AU} activations. Both streams employ capsule-based computations with the extracted individual primary capsules combined and \textit{routed} together to evaluate their influence on final class-capsules. The class-capsules are also passed to a decoder that learns to reconstruct the middle \ac{FoI}, further regularising learning.
%  For a given input sequence window, the model learns to selectively focus on relevant spatial and spatio-temporal features.

%  given what stage of the \ac{AU} lifecycle is represented in the input sequence.

\subsection{Windowed Video Sequences as Input}
\ac{AULA-Caps} takes as input a video sequence consisting of contiguous ($96 \times 96$) frames of normalised face-centred images (each pixel $p\in[-1,1]$). Input sequences are generated by taking each frame of the video, along with $N$ frames immediately preceding and succeeding the frame. The middle \ac{FoI} is passed to the spatial processing stream while the entire window of $2N+1$ frames is processed using the spatio-temporal processing stream. The overall task for the model is to predict the activated \acp{AU} in the \ac{FoI}. Here, we set $N$=$2$ forming an input window of $5$ frames.

% For both the datasets, we first extract individual frames from the video sequences and extract face-centred images using the facial landmark information provided. Whenever facial landmarks are not provided, the dlib\footnote{\tiny\url{http://dlib.net}} Python Library is used to extract face-centred images.

\begin{figure}[t]
\centering

\begin{subfigure}{0.4\textwidth}
  \includegraphics[width=\textwidth]{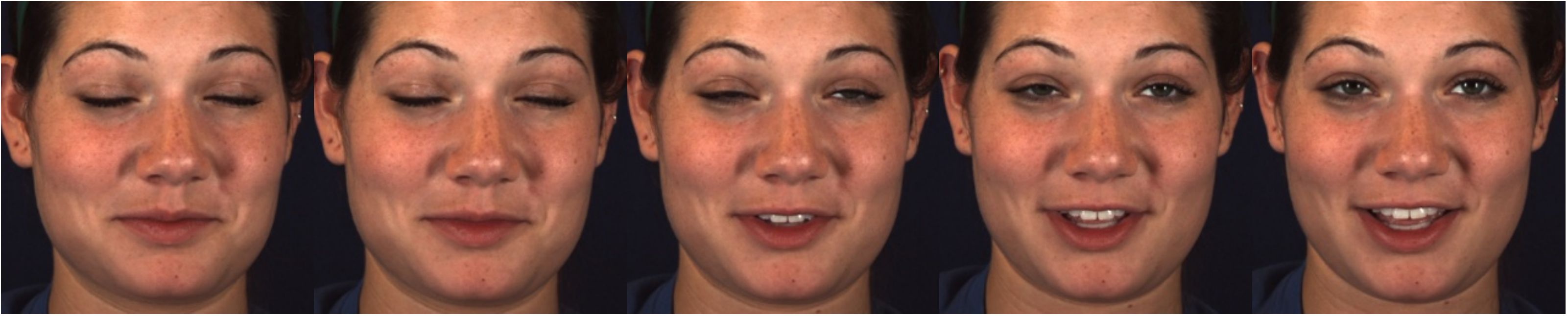}
\caption{Onset Segment sample from BP4D.} \label{onset}
%   \vspace{3mm}
\end{subfigure}\\%\hfil % <-- added
\begin{subfigure}{0.4\textwidth}
  \includegraphics[width=\textwidth]{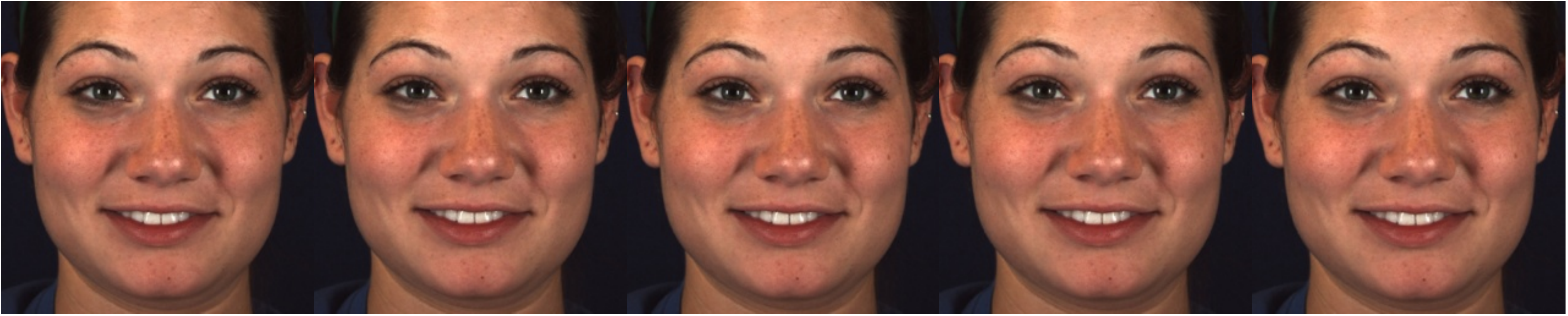}
  \caption{Apex Segment sample from BP4D.}
  \label{apex}
\vspace{-2mm}

\end{subfigure}
\caption{Onset and Apex segment contiguous frames.}
\label{fig:lifecycle-aware}
\vspace{-4mm}
\end{figure}

\subsection{Motivation for Lifecycle-Awareness}
Following the \ac{AU} lifecycle, different segments of activation, that is, \textit{onset}, \textit{apex} and \textit{offset} form the evolution of an \ac{AU}. In the \textit{onset} and \textit{offset} phases, the input images, or the extracted facial features, have high \textit{variation}, in that, the contiguous frames are sufficiently different, as illustrated in Figure~\ref{onset}. Thus, focusing on this difference provides important temporal information about \ac{AU} activation patterns. In \textit{apex} segment frames, however, the contiguous frames have low \textit{variation} and are not sufficiently different, as illustrated in Figure~\ref{apex}. As a result, instead of focusing on temporal changes, spatial features extracted from just a single \ac{FoI} provide sufficient information for \ac{AU} prediction. 

The two streams of processing in the \ac{AULA-Caps} model are designed to exploit this difference by extracting relevant spatial and spatio-temporal features and combining them in a manner where their individual contribution is weighted based on their relevance for \ac{AU} prediction. 
% It is expected that for \textit{off-peak} input windowed-sequences, the model will focus more on spatio-temporal features while for \textit{near-apex} sequences, spatial features will take precedence. 
Realising such an ability for the model to selectively and automatically tune into separate features based on where in the \ac{AU} lifecycle the input sequence originates from, motivates the \textit{lifecycle-awareness} of the \ac{AULA-Caps} model.

% \subsection{Multi-stream Capsule}

% At each instance, the model not only looks at the frame-of-interest using the spatial stream, but also considers the immediately preceding and succeeding frames, over a window, to learn temporal dependencies between features from these contiguous frames. This is beneficial to be sensitive to how \acp{AU} evolve through their \textit{lifecycle}. The model learns to focus on spatial as well as spatio-temporal features, given their individual contribution towards \ac{AU} detection. We hypothesis that sequence windows near the \textit{apex} frames in the lifecycle of the an \ac{AU} will result in very different activation signatures for the spatio-temporal primitives as compared to windows belonging to the \textit{onset} or \textit{offset} segments. This is because within each window, the feature activations will be higher for these segments as the difference between contiguous frames will be significant compared to \textit{apex} frames with the model standing to gain more from learning spatio-temporal dynamics between these samples.  During peak intensity, contiguous frames do not vary a lot and it may be more beneficial for the model to focus only on spatial information. On the other hand, during \textit{neutral}-to-\textit{onset} or \textit{offset}-to-\textit{neutral} transitions, focusing on the temporal changes may be more relevant.
% \textit{(to be shown in visualisations of 3d stream)}

\subsection{Computing Spatial Features}
The spatial processing stream (see Spatial Stream in Figure~\ref{fig:aula_model}) takes as input the middle \ac{FoI} (f) from an input sequence and passes it through a convolutional (conv) layer with $128$ filters of size ($7\times 7$) followed by \textit{BatchNormalisation} and a \textit{LeakyReLU} ($\alpha=0.2$) activation. The output of this conv layer is passed through two Residual blocks consisting of multi-resolution conv layers with shortcut connections~\cite{He2016ResNet}, with $128$ and $64$ filters for each conv layer in the respective blocks using a \textit{LeakyReLU} ($\alpha=0.2$) activation. Each residual block is followed by a ($2\times2$) maxpooling layer for dimensionality reduction. The output of the final maxpooling layer is passed to the Primary Capsule layer consisting of a conv layer followed by reshaping and squashing of the extracted spatial features into $576$ capsules of $16$ dimensions each. 

% The 2D-stream processes the middle frame (f) from the input sequence using a convolutional (conv) layer with $5$ filters to extract features from each individual frame. This results in $5$ feature maps for each of the frames in the input sequence window. These are then stacked together as streams and this input is then passed through $3$ stacked conv layers (with $64, 32, 32$ filters of size ($5\times 5$) each with a stride of $2$ in each dimension) with \textit{ReLU} activation, followed by \textit{BatchNormalisation}. The output of the final conv layer is passed to a Primary Capsule layer (PrimeCaps) consisting of another conv layer with linear activation followed by squashing of the extracted 2D features into $48$ capsules of $16$ dimensions each. 

\subsection{Computing Spatio-Temporal Features}

The spatio-temporal processing stream (see Spatio-temporal Stream in Figure~\ref{fig:aula_model}) processes the entire input window sequence. The sequence is first passed through a $3$DConv layer consisting of $128$ filters of size ($5\times5\times5$) followed by \textit{BatchNormalisation} and a \textit{LeakyReLU} ($\alpha=0.2$) activation. The output of the $3$DConv layer is passed through two $3$DConv blocks consisting of two conv layers each followed by \textit{BatchNormalisation} and \textit{LeakyReLU} ($\alpha=0.2$) activation. Conv layers in the first block consist of $128$ filters of size ($5\times5\times5$) while the second block layers consist of $64$ filters. Each block is followed by a ($2\times2\times2$) $3$D maxpooling layer for dimensionality reduction. The final maxpooling layer output is passed to the $3$D Primary Capsule layer consisting of a $3$DConv layer followed by reshaping and squashing of the extracted spatio-temporal features into $864$ capsules of $16$ dimensions each. 
% The 3D stream processes the input sequences using $3$ stacked 3DConv layers (with $64, 128, 256$ filters each of size ($2\times 5\times 5$) with a stride of $2$) with \textit{ReLU} activation, followed by \textit{BatchNormalisation}. The output of the final 3D Conv layer is reshaped and passed to a Primary Capsule layer (PrimeCaps), squashing the extracted 3D features into $48$ capsules of $16$ dimensions each. 

\subsection{Combining Extracted Features}

The extracted \textit{primary} capsules representing spatial and spatio-temporal primitives computed by the two streams are concatenated together resulting in $1440$ capsules of $16$ dimensions each. The iterative \textit{routing-by-agreement} algorithm~\cite{CAPS2017} then couples the concatenated capsules with the \ac{AU} Capsule Layer (\ac{AU}-Caps), computing $12$ capsules each corresponding to an \ac{AU} label (see Table~\ref{tab:AU_labels}). The output of the \ac{AU}-Caps layer is used to predict the \acp{AU} activated in the \ac{FoI}, replacing the computed capsule with its length \textit{squashed} between $[0,1]$ depicting the activation probability for each \ac{AU} label. The \ac{AU}-Caps layer output is also used by the Decoder model to reconstruct the \ac{FoI}.

\begin{table}[t]
    \centering
    \rowcolors{2}{gray!10}{white}
    \caption{Action Units examined in this work.}
    \label{tab:AU_labels}
    \vspace{-2mm}
    {
    \scriptsize
    \setlength\tabcolsep{3.8pt}    % \setlength\tabcolsep{2.0pt}
    \begin{tabular}{cl|cl|cl}\toprule
        \textbf{\ac{AU}} & \makecell[c]{\textbf{Description}} & \textbf{\ac{AU}} & \makecell[c]{\textbf{Description}} & \textbf{\ac{AU}} & \makecell[c]{\textbf{Description}}                \\\midrule
         \textbf{1}  & Inner Brow Raiser &  \textbf{7}  & Eyelid Tightener      &   \textbf{15}  & Lip Corner Depressor \\\midrule
         \textbf{2}  & Outer Brow Raiser &  \textbf{10}  & Upper Lip Raiser     &   \textbf{17}  & Chin Raiser          \\\midrule
         \textbf{4}  & Brow Lowerer      &  \textbf{12}  & Lip Corner Puller    &   \textbf{23}  & Lip Tightener        \\\midrule
         \textbf{6}  & Cheek Raiser      &  \textbf{14}  & Dimpler              &   \textbf{24}  & Lip Pressor          \\\bottomrule
    \end{tabular}
    }
        \vspace{-4mm}

\end{table}

\subsection{Decoder for Image Reconstruction}
The Decoder is used to regularise learning in the model~\cite{CAPS2017} making sure it learns task-relevant features, as well as to enable visualisation of learnt features through the reconstructed images. The \ac{AU}-capsules are masked using the label \textit{y} for reconstructing the \ac{FoI} ($x_{gen}$). In \acs{AULA-Caps}, we use transposed conv layers for the decoder, instead of dense layers proposed by Sabour \etal~\cite{CAPS2017}. This significantly reduces the number of parameters in the decoder ($\approx2.8$M as opposed to $10$M in~\cite{CAPS2017}) while also improving the photo-realistic quality of reconstructed images. The decoder, adapted from the generator of~\cite{Churamani2020CLIFER}, implements $4$ stacked conv layers, using \textit{ReLU} activation, performing transposed convolutions with $128, 64, 32, 16$ filters, respectively, of size ($5\times5$) each with a stride of ($2\times2$). The output of the last conv layer is passed through another transposed conv layer using \textit{tanh} activation to generate the resultant image $(x_{gen})$ with the same dimensions as the \ac{FoI} ($x_{r}$). 

% An $\mathcal{L}_1$ \textit{reconstruction loss} ($\mathcal{L}_{rec}$) is imposed on $G$, enabling reconstruction of images:

\subsection{Learning Objectives}
The two streams of \ac{AULA-Caps}, along with the decoder, are trained together in an end-to-end manner. The \ac{AULA-Caps} model predicts $2$ outputs in each run, that is, the activation probabilities for $12$ \acsp{AU} (see Table~\ref{tab:AU_labels}) as well as the reconstructed \ac{FoI}. The learning objectives for the model are as follows:

% The two-streamed \ac{AULA-Caps} network for \ac{AU} detection along with the decoder model for reconstructing images are trained together in an end-to-end fashion. The learning objectives for the models are described as:
\vspace{-1mm}
\paragraph{\ac{AU} Prediction:}
The \ac{AULA-Caps} model predicts the activation probabilities for the $12$ \acsp{AU} in the \ac{FoI} as the length of the \ac{AU}-class capsules. Learning to detect the activated \acp{AU} focuses on minimising a \textit{weighted} margin loss. The loss for each of the \acp{AU} ($\mathcal{L}_{au}$) is defined as:
{\small
\begin{equation}
    \begin{aligned}
        \mathcal{L}_{au} = w_{au} (&T_{au}\max(0, m^{+} - ||p_{au}||)^2 \\
        & + \lambda_{au} (1-T_{au})\max(0, ||p_{au}|| - m^{-})^2),
        \label{eq:Lmargin}
    \end{aligned}
\end{equation}
}{\noindent}where $T_{au}=1$ if an \ac{AU} is present and $0$ otherwise, $||p_{au}||$ is the prediction (output probability) for an \ac{AU} computed as the magnitude (length) of the respective class capsule, $m^+$ and $m^-$ are the positive and negative sample margins, $\lambda_{au}$ is a constant weighting the effect of positive and negative samples, and $w_{au}$ is a class balancing weight. We set $m^+ = 0.9$, $m^- = 0.1$ and $\lambda_{au} = 0.5$, following~\cite{CAPS2017}. $w_{au}$ is computed using the occurrence-rate for the respective \acp{AU} in the training data. This is done to reduce the effect of the class imbalance under multi-label classification settings. Following~\cite{Shao2018Deep}, $w_{au}$ is computed as follows:
{\small
\begin{equation}
    \begin{aligned}
       w_{au} = \frac{(1/r_i)N}{\sum_i^N (1/r_i)},
        \label{eq:Lmargin}
    \end{aligned}
\end{equation}
}where $N$ is the number of \acsp{AU} (in this case $N=12$) and $r_i$ is the occurrence rate of \ac{AU}$_i$. The resultant loss ($\mathcal{L}_{margin}$) is computed as the sum of the losses for each \ac{AU} ($\mathcal{L}_{au}$).

% {
% % \small
% \begin{equation}
%     \begin{aligned}
%       \mathcal{L}_{margin} = \sum \mathcal{L}_{au}.
%     \end{aligned}
% \end{equation}
% }
\vspace{-1mm}
\paragraph{Image Reconstruction:}
The Decoder reconstructs the \ac{FoI} using the extracted \ac{AU} capsules imposing a mean squared error \textit{reconstruction loss} ($\mathcal{L}_{rec}$):

{\small
\begin{equation}
    \begin{aligned}
    \min_{X_r, X_{gen}}\mathcal{L}_{rec} = L_2(x_r, x_{gen}),
    \end{aligned}
\end{equation}
}where $x_r$ is the \ac{FoI} and $x_{gen}$ is the reconstructed image. 

% Furthermore, a pre-trained VGG-face model~\cite{Parkhi15} is used for ID preservation (similar to~\cite{ding2017exprgan,Churamani2020CLIFER}), evaluating the facial features extracted from the input and generated images. $L_{ID}$ is computed as:

% {
% % \small
% \begin{equation}
%      \min_{X_r, X_{gen}}\mathcal{L}_{ID} = \sum_{l}L_1(\phi_l(x_{gen}),\phi_l(x_r)),
%     \label{eq:Lid}
%     % %\vspace*{-2mm}
% \end{equation}
% }where $\phi_l$ represents the $l$-th layer VGG-face features. The activation from the first $5$ convolution layers are compared for input and generated images. 
% Additionally, a total variation regularisation ($L_{tv}$)~\cite{Mahendran_2015_CVPR} is imposed (similar to~\cite{ding2017exprgan}) that uses the sum of the absolute differences for neighbouring pixel-values in the input images to avoid ‘ghosting’ artefacts.

% The decoder loss is a weighted combination of these two losses and is given as:
% {
% % \small
% \begin{equation}
%     \mathcal{L}_{Decoder} = \lambda_1 \mathcal{L}_{rec} + \lambda_2 \mathcal{L}_{ID} %+ \lambda_3 \mathcal{L}_{tv}
%     \label{eq:decoder}
% \end{equation}
% }Following~\cite{ding2017exprgan,Churamani2020CLIFER}, the weights for the different loss terms are set to $\lambda_1=1$, $\lambda_2=0.33$
%, $\lambda_3=8.5\times10^{-5}$

\paragraph{Overall Objective:}
The overall objective for \ac{AULA-Caps} is a weighted sum of the overall \ac{AU} prediction ($\mathcal{L}_{margin}$) and image reconstruction ($\mathcal{L}_{rec}$) objectives:
{
\small
\begin{equation}
    \begin{aligned}
        \mathcal{L}_{AULA} = \mathcal{L}_{margin} + \lambda_{d}\mathcal{L}_{rec},
    \end{aligned}
\end{equation}
}where $\lambda_{d}$ is set to $0.05$ balance the loss terms. 

%-------------------------------------------------------------------------
%-------------------------------------------------------------------------

\section{Experiments}
% We train and test \ac{AULA-Caps} on two popular \ac{AU} prediction datasets. This sections describes the experimental settings and discusses the results obtained:

\subsection{Datasets}

We train and test \ac{AULA-Caps} on two popular \ac{AU} benchmark datasets, namely GFT and BP4D. For both datasets, samples representing the $12$ most frequently occurring \acp{AU} are used (see Table~\ref{tab:AU_labels}).

\vspace*{-2mm}
\paragraph{GFT:} The Sayett \acf{GFT} Dataset~\cite{GFTDatabase} consists of $1$-minute video recordings from $96$ subjects, spontaneously interacting with each other in group settings ($2-3$ persons per group). The interactions are unstructured, allowing for natural and spontaneous reactions by the participants, annotated for each group-member at frame-level. Annotations are made available at frame-level.

\vspace*{-2mm}
\paragraph{BP4D:} The \acs{BP4D} dataset~\cite{ZHANG2014BP4D} consists of video sequences from $41$ subjects performing $8$ different affective tasks to elicit emotional reactions. Approximately $500$ frames for each video are annotated for occurrence and intensity of the activated \acsp{AU}. In our experiments, we only use occurrence labels for \ac{AU} detection.
\newline

% \vspace*{1mm}
{\noindent}Both BP4D and GFT represent different data settings, enabling a comprehensive evaluation of the proposed model. While GFT represents complex, naturalistic recording settings, BP4D, on the other hand, consists of cleaner, face-centred images and provides much more data per subject.

\subsection{Experiment Settings}
\paragraph{Evaluation Metric:}
Similar to other approaches~\cite{Ertugrul2019Cross,nott44740,shao2020spatiotemporal}, we follow $3-$fold cross-validation for our evaluations, splitting the data into $3$ folds where each subject occurs in test-set once. For each run, the model is trained on $2$ folds and tested on the third. Results are collated across the $3$ folds. We report model performance using \textit{F1-Scores} computed as the harmonic mean  (F1$=\frac{2RP}{R+P}$) of the precision ($P$) and recall ($R$) scores, providing for a robust evaluation of the model. F1-score is the most commonly employed metric for reporting \ac{AU} detection performance~\cite{Valstar2011FERA}.

% \vspace*{-1mm}
\paragraph{Implementation Details:}
The \ac{AULA-Caps} model is implemented using Keras-Tensorflow\footnote{\tiny\url{https://www.tensorflow.org/versions/r1.15/api_docs/python/tf/keras}}. We train the model individually on each dataset in an end-to-end manner using the \textit{Adam} optimiser with an initial learning rate of $2.0e^{-4}$ and decayed each epoch by a factor of $0.9$. For each fold, the model is trained for $12$ epochs with early stopping with a batch-size of $24$. The model hyper-parameters namely, the filter sizes, number of filters for each layer, capsule dimension size, the batch-size and the initial learning-rate are optimised using the Hyperopt\footnote{\tiny\url{http://hyperopt.github.io/hyperopt/}} Python Library. The models are trained on a NVIDIA GeForce GTX 1080Ti GPU.

\subsection{Results}
% Both the BP4D and GFT datasets represent different data settings enabling a comprehensive evaluation of the proposed model. While GFT represents naturalistic and complex recording settings, BP4D on the other hand consists of cleaner, face-centred images and provides much more data samples per subject. 
% Here we discuss the results for the \ac{AULA-Caps} model on the two datasets in comparison with state-of-the-art approaches in literature.

% \vspace*{-1mm}
\paragraph{GFT:}

% The GFT dataset is composed of video samples representing complex evaluation settings particularly as subjects are recorded naturalistically interacting with each other in group settings. Furthermore, the subjects perform a 'drink-tasting' task~\cite{GFTDatabase} resulting a lot of the frames containing occlusions and varying perspectives. These constraints result in a huge imbalance in the data with respect to the different \ac{AU} annotations.

% applying region-based convolutional computations that capture local appearance changes for different facial regions [JPML]~\cite{Zhao2015JPML},
Table~\ref{tab:performance-metrics-GFT} presents the model performance results on the GFT dataset in comparison to the state-of-the-art. We compare \ac{AULA-Caps} to different spatial and spatio-temporal approaches such as the \acs{CNN}-based cross-domain learning [CRD]~\cite{Ertugrul2019Cross}, an Alex-Net-based model [ANet] for frame-based \ac{AU} detection~\cite{Chu2017Learning}, the [J$\hat{A}$A]~\cite{Shao2020} approach that uses multi-scale high-level facial features extracted from face alignment tasks to aid \ac{AU} prediction, and learning temporal variation in facial features using a CNN-LSTM model~\cite{Chu2017Learning}. The [CNN-LSTM] model applies frame-based spatial computations and extends this learning to the temporal domain by evaluating how spatial features evolve over time. In contrast, the proposed \ac{AULA-Caps} model simultaneously extracts spatial and spatio-temporal features from input sequences and learns to combine them to selectively focus on relevant features for respective \ac{AU} predictions.

\ac{AULA-Caps} achieves the best results for $4$ \ac{AU} labels and second-best results for another $3$. Despite achieving the second-best overall results (-$0.002$ difference in \textit{Avg.} F1-score from the best approach LSTM), the model performs rather poorly for \ac{AU} $1, 4$ and $14$ impacting the overall score of the model. We discuss these results in Section~\ref{sec:performance-discussion}.
% Certain \acsp{AU} such as \ac{AU} $1, 4, 14$ contain much less samples than the others resulting in the poor performance of the model for these \acsp{AU}. Nonetheless, overall, for the $12$ \ac{AU} classes, the proposed model achieves competitive results to the state-of-the-art approaches achieving second-best results amongst the compared approaches. %A discussion on the model is presented in Section~\ref{sec:discuss_lifecycle}.

% \subsubsection{Data Pre-processing}
% For both the datasets, we first extract individual frames from the video sequences and extract face-centred images using the facial landmark information provided. Whenever facial landmarks are not provided, the dlib\footnote{\tiny\url{http://dlib.net}} Python Library is used to extract face-centred images. Each image is then resized into $96\times96$ grayscale frames and normalised such that each pixel value is scaled $\in[-1,1]$.

% The BP4D dataset on the other hands represents controlled recording settings with where the subjects are asked to perform certain affective tasks that are designed to elicit specific affective responses. The subjects are recorded from various perspectives providing for a lot variation in subject-specific data. The class-imbalance problem is also seen in the BP4D dataset with $4$ out of the $12$ \acsp{AU} dominating the data distribution.

% \vspace*{-1mm}
\paragraph{BP4D: }
Table~\ref{tab:performance-metrics-BP4D} presents the \ac{AULA-Caps} results for BP4D and compares them to the state-of-the-art approaches such as the [CNN-LSTM]~\cite{Chu2017Learning} learning temporal variation in facial features, the [EAC] method~\cite{Li2018EACNet} that employs enhancing and cropping mechanism to focus on selective regions in an image, the [ROI] network~\cite{Li2017ROI} that focuses on learning regional features using separate local \acs{CNN}, a 2D Capule-Net based model [CapsNet] proposed by~\cite{rashid2018facial}, the [J$\hat{A}$A]~\cite{Shao2020} approach that uses multi-scale high-level facial features, the semantic learning-based [SRERL]~\cite{li2019semantic} model and the [STRAL]~\cite{shao2020spatiotemporal} approach that employs a spatio-temporal graph convolutional network to capture both spatial and temporal relations for \ac{AU} prediction. The \ac{AULA-Caps} model, on the other hand, uses a multi-stream approach that simultaneously learns and combines spatial as well as spatio-temporal features making it sensitive to the temporal evolution of \ac{AU} activations. 

\ac{AULA-Caps} achieves the best results for $3$ \ac{AU} labels and second-best results for another $3$. Overall, the model outperforms other models, with closest \textit{Avg.} F1-score difference to the STRAL approach~\cite{shao2020spatiotemporal} being $0.013$ with both STRAL and the \ac{AULA-Caps} model combining spatial and spatio-temporal analysis of facial features.

\begin{table}[t]
    \centering
    \rowcolors{3}{gray!10}{white}
    \setlength\tabcolsep{4.0pt}
    \caption{Performance Evaluation (F1-Scores) on GFT Dataset: %JPML~\cite{Chu2017Learning, Zhao2015JPML}, 
    CRD~\cite{Ertugrul2019Cross}, 
    ANet~\cite{Chu2017Learning},
    J$\hat{A}$A~\cite{Shao2020},
    CNN-LSTM~\cite{Chu2017Learning}. \textbf{Bold} values denote best while [\textit{bracketed}] denote second-best values for each row. All scores reported from the respective papers. \tt{\small$^*$Averaged for $10$ \acsp{AU}.}}
    \label{tab:performance-metrics-GFT}
    {
    \scriptsize
    \begin{tabular}{c|cccccc}%cccc|}
    \toprule
    % \multirow{2}{*}{\textbf{AU}}  & \multicolumn{6}{c}{\textbf{F1-Score }} \\ \cmidrule{2-7}  %& \multicolumn{4}{c|}{\textbf{AUC-Score}} \\ \cline{2-13} 
     \textbf{AU} 
     %& \textbf{JPML} %& \textbf{TCAE} 
     & \textbf{CRD} & \textbf{ANet} & \textbf{J$\hat{A}$A} & \textbf{CNN-LSTM}  & \textbf{\ac{AULA-Caps}} \\ \midrule %& \textbf{[1]} & \textbf{[2]} & \textbf{CRD} & \multicolumn{1}{l|}{\textbf{Ours}} \\ \midrule
    1  %& 0.175 %& [\textit{0.439}] 
    &  [\textit{0.437}] & 0.312 & \textbf{0.465} & 0.299 & 0.313\\ \midrule %& & -- & -- & 0.953 & -- \\ \midrule
    2  %& 0.209 %& [\textit{0.495}] 
    & 0.449 & 0.292 & [\textit{0.493}] & 0.257  & \textbf{0.498}\\ \midrule %& & -- & -- & 0.935 & -- \\\midrule
    4  %& 0.032 %& 0.063 
    & 0.198 & \textbf{0.719} & 0.192 & [\textit{0.689}] & 0.297\\ \midrule %& & -- & -- & 0.982 & -- \\\midrule
    6 % & 0.705 %& 0.710 
    & 0.746 & 0.645 & \textbf{0.790} & 0.673 & [\textit{0.775}]\\ \midrule %& & -- & -- & 0.882 & -- \\\midrule
    7  %& 0.655 %& -- 
    & 0.721 & 0.671 & -- & [\textit{0.725}] & \textbf{0.772}\\ \midrule %& & -- & -- & 0.791 & -- \\\midr1ule
    10  %& 0.679 %& [\textit{0.762}]  
    & \textbf{0.765} & 0.426 & [\textit{0.75}] & 0.670  & 0.749\\ \midrule %& & -- & -- & 0.840 & -- \\\midrule
    12  %& 0.742 %& 0.795 
    & [\textit{0.798}] & 0.731 & \textbf{0.848} & 0.751 & 0.785\\ \midrule %& & -- & -- & 0.905 & -- \\\midrule
    14  %& 0.524 %& 0.107 
    & 0.500 & [\textit{0.691}] & 0.441 & \textbf{0.807} & 0.236\\ \midrule %& & -- & -- & 0.691 & -- \\\midrule
    15  %& 0.203 %& 0.285 
    & 0.339 & 0.279 & 0.335 & \textbf{0.435} & [\textit{0.371}]\\ \midrule %& & -- & -- & 0.875 & -- \\\midrule
    17  %& 0.483 %& -- 
    & 0.170  & [\textit{0.504}] & -- & 0.491 & \textbf{0.592}\\ \midrule %& & -- & -- & 0.898 & -- \\\midrule
    23  %& 0.318 %& 0.345 
    & 0.168  & 0.348 & \textbf{0.549} & 0.350 & [\textit{0.522}]\\ \midrule %& & -- & -- & 0.928 & -- \\\midrule
    24  %& 0.285 %& 0.417 
    & 0.129 & 0.390 & [\textit{0.507}] & 0.319 & \textbf{0.530}\\ \midrule %& & -- & -- & 0.962 & -- \\ \midrule
    \textbf{Avg.}  %& 0.418  %& 0.442$^*$ 
    & 0.452 & 0.500 & 0.537$^*$ & \textbf{0.539} & [\textit{0.537}] \\ \bottomrule 
    \end{tabular}%
    }
    % \vspace{-3mm}

\end{table}

\begin{table}[t]
    \centering
    \vspace{4mm}
    \rowcolors{2}{gray!10}{white}
    \caption{Ablations on \ac{AULA-Caps} for BP4D. Decoder parameters ($\approx2$M) excluded for comparison with CNN baselines. Batch size is set to $24$.}
    \label{tab:performance-metrics-ablation-channels}
    {
    \scriptsize
    \begin{tabular}{lccc}
    \toprule
    \makecell[c]{\textbf{Model}} & \textbf{Avg. F1-Score} & \textbf{\#Params} & \textbf{RunTime / Batch} \\\midrule
    
    2D CNN Baseline & 0.573 & 3.44M & 0.31s \\\midrule
    3D CNN Baseline & 0.540 & 15.09M &  0.63s \\\midrule
    % Multi-stream CNN Baseline & \\\midrule
    2D Stream \ac{AULA-Caps}  & 0.580 & 3.06M   & 0.35s \\\midrule
    3D Stream \ac{AULA-Caps}  & 0.550 & 8.46M   & 0.66s \\\midrule
    \ac{AULA-Caps}  & \textbf{0.645}  & 11.51M  & 1.22s \\\bottomrule
    
    % 2D CNN Baseline & 0.573 & 3.44M & -- \\\midrule
    % 3D CNN Baseline & 0.540 & 15.09M  & --\\\midrule
    % % Multi-stream CNN Baseline & \\\midrule
    % 2D Stream \ac{AULA-Caps}  & 0.580 & 3.02M   & 0.62s \\\midrule
    % 3D Stream \ac{AULA-Caps}  & 0.550 & 8.42M   & 0.95s \\\midrule
    % \ac{AULA-Caps}  & \textbf{0.537}  & 12.41M  & 1.11s \\\bottomrule
    \end{tabular}%
    }
    % \vspace*{-3mm}

\end{table}

\subsection{Ablation: Spatial \vs Spatio-Temporal Features}
\label{sec:ablations}
% As the proposed model is composed of different components and design choices, it is important to evaluate how each of these influence the model performance. Two main aspects that form the basis of our approach is processing windows of input sequences rather than performing frame-based analysis and extracting spatial and spatio-temporal features separately using the two-channels of the model. Thus, to understand how these influence model performance, we conduct the following ablation studies:

% ADD 2D and 3D CNN implementations (along with combined model) 
% \subsubsection{Spatial \vs Spatio-Temporal Features}
Since \ac{AULA-Caps} focuses on learning spatial and spatio-temporal features simultaneously, it is important to understand how each of these feature-sets contribute towards the overall performance of the model. To evaluate the contribution of the learnt spatial features, we use the trained $2$D stream layers to predict \acp{AU} by appending a separate \ac{AU}-Caps layer to the primary capsule layer. The weights of the $2$D stream are frozen and only the routing algorithm is run for the added \ac{AU} capsule layer. Similarly, for assessing the effect of learning spatio-temporal dependencies across contiguous frames, we use the trained spatio-temporal ($3$D) stream layers to predict the \ac{AU} labels by appending a separate \ac{AU}-Caps layer to the primary capsules.

Furthermore, for highlighting the contribution of capsule-based computation, we compare these results with $2$D and $3$D \ac{CNN}-based models. The two streams are unchanged with only the capsule-block replaced by fully-connected dense layers. The results on BP4D for the different ablations conducted are presented in Table~\ref{tab:performance-metrics-ablation-channels}. Analysing the ablations on the BP4D provides a fairer comparison as it consists of more samples per subject with cleaner and face-centred images.

% \subsubsection{Setting the Window-size}
% The proposed \ac{AULA-Caps} model evaluates a window of input video sequence to predict the activated \acsp{AU} in the middle \ac{FoI}. This provides context about the temporal dynamics of \ac{AU} activation represented in the window. It is important, therefore, to evaluate how deciding on a window-size affects the model performance. For this, we choose $3$ different window-sizes taking into account $1, 2$ and $3$ neighbouring frames in each direction and compare the model performance on these window-sizes for the BP4D dataset. The results obtained are presented in Table~\ref{}. 

\begin{table}[t]
    \centering
    \rowcolors{3}{gray!10}{white}

    \setlength\tabcolsep{2.0pt}
    \caption{Performance Evaluation (F1-Scores) on BP4D Dataset: CNN-LSTM~\cite{Chu2017Learning}, EAC~\cite{Li2018EACNet}, 
    ROI~\cite{Li2017ROI},
    CapsNet~\cite{rashid2018facial},
    J$\hat{A}$A~\cite{Shao2020}, 
    SRERL~\cite{li2019semantic}, STRAL~\cite{shao2020spatiotemporal}. \textbf{Bold} values denote best while [\textit{bracketed}] denote second-best values for each row. All scores reported from the respective papers.}
    \label{tab:performance-metrics-BP4D}
    {
    \scriptsize
    \hspace*{-2mm}
    \begin{tabular}{c|cccccccc}
    \toprule
    % \multirow{2}{*}{\textbf{AU}} & 
    % \multicolumn{4}{c|}{\textbf{Accuracy}} & 
    % \multicolumn{8}{c}{\textbf{F1-Score}} \\ \cmidrule{2-9} 
    % & \textbf{EAC} & \textbf{J$\hat{A}$A} & \textbf{STRAL} & \multicolumn{1}{c}{\textbf{Ours}} & 
    \textbf{AU} & \textbf{CNN-LSTM} & \textbf{EAC} %& \textbf{TCAE} 
    & \textbf{ROI} & \textbf{CapsNet} & \textbf{J$\hat{A}$A} & \textbf{SRERL} & \textbf{STRAL}  & \multicolumn{1}{c}{\textbf{\ac{AULA-Caps}}} \\ \midrule
    
    1 &
    % 0.689 & 0.752 & [\textit{0.776}] &  \multicolumn{1}{c|}{\textbf{0.846}} & 
    0.314  & 0.390 %& 0.431 
    & 0.362  & 0.468 & [\textit{0.538}] & 0.469 & 0.482 & 
    \multicolumn{1}{c}{\textbf{0.562}}  \\ \midrule
    % \multicolumn{1}{c}{\textbf{0.602}}  \\ \midrule
    
    2 & 
    % 0.739 & [\textit{0.802}] & 0.781 & \multicolumn{1}{c|}{\textbf{0.860}} & 
    0.311 & 0.352 %& 0.322 
    & 0.316 & 0.291  & \textbf{0.478} & 0.453 & [\textit{0.477}] &
    \multicolumn{1}{c}{0.465} \\\midrule
    % \multicolumn{1}{c}{\textbf{0.490}} \\\midrule
    
    4 & 
    % 0.781 & [\textit{0.829}] & 0.824 & \multicolumn{1}{c|}{\textbf{0.869}} & 
    \textbf{0.714} & 0.486 %& 0.444 
    & 0.434 & 0.529 & [\textit{0.582}] & 0.556 & 0.581 & \multicolumn{1}{c}{0.573} \\\midrule
    % \multicolumn{1}{c}{[\textit{0.617}]} \\\midrule
    
    6 & 
    % 0.785  & [\textit{0.798}] & 0.776 & \multicolumn{1}{c|}{\textbf{0.820}} & 
    0.633 & 0.761 %& 0.751 
    & 0.771 & 0.753 & [\textit{0.785}] & 0.771 & 0.758 & \multicolumn{1}{c}{\textbf{0.796}} \\\midrule
    % \multicolumn{1}{c}{\textbf{0.827}} \\\midrule
    
    7 & 
    % 0.69 & 0.723 & \textbf{0.764} &  \multicolumn{1}{c|}{[\textit{0.759}]} & 
    0.771 & 0.729 %& 0.705 
    & 0.737 & 0.776 & 0.758 & \textbf{0.784} & [\textit{0.781}] &  \multicolumn{1}{c}{0.765}  \\\midrule
    % \multicolumn{1}{c}{\textbf{0.798}}  \\\midrule
    
    10 & 
    % 0.776 & [\textit{0.782}] & [\textit{0.782}] &  \multicolumn{1}{c|}{\textbf{0.828}} &
    0.450 & 0.819 %& 0.808 
    & \textbf{0.850} & 0.824 & 0.827 & 0.835 & 0.816 &  \multicolumn{1}{c}{[\textit{0.843}]}\\\midrule
    % \multicolumn{1}{c}{\textbf{0.861}}\\\midrule
    
    12 & 
    % 0.846 & \textbf{0.866}  & [\textit{0.858}] & \multicolumn{1}{c|}{0.852} & 
    0.826 & 0.862 %& 0.855 
    & 0.870 & 0.850 & \textbf{0.882}  & [\textit{0.876}] & [\textit{0.876}] & \multicolumn{1}{c}{0.874}  \\\midrule
    % \multicolumn{1}{c}{0.867}  \\\midrule
    
    14 & 
    % 0.606 & [\textit{0.651}]  & 0.636  & \multicolumn{1}{c|}{\textbf{0.734}} & 
    \textbf{0.729} & 0.588 %& 0.618 
    & 0.626 & 0.657 & 0.637 & 0.639 & 0.605 & \multicolumn{1}{c}{[\textit{0.718}]} \\\midrule
    % \multicolumn{1}{c}{\textbf{0.746}} \\\midrule
    
    15 & 
    % 0.781 & 0.810 & \textbf{0.843}  & \multicolumn{1}{c|}{[\textit{0.839}]} & 
    0.340 & 0.375 %& 0.347 
    & 0.457 & 0.337 & 0.433 & \textbf{0.522} & [\textit{0.502}] &  \multicolumn{1}{c}{0.457}   \\\midrule
    % \multicolumn{1}{c}{0.454}   \\\midrule
    
    17 & 
    % 0.706 & [\textit{0.728}] & 0.716 & \multicolumn{1}{c|}{\textbf{0.732}} & 
    0.539 & 0.591 %& 0.585 
    & 0.580 & 0.606 & 0.618 & 0.639 & [\textit{0.640}] & \multicolumn{1}{c}{\textbf{0.694}}  \\\midrule
    % \multicolumn{1}{c}{\textbf{0.690}}  \\\midrule
    
    23 &
    % 0.81 & [\textit{0.829}] & 0.823 & \multicolumn{1}{c|}{\textbf{0.861}} & 
    0.386 & 0.359 %& 0.372 
    & 0.383 & 0.369 & 0.456 & 0.471 & \textbf{0.512} & \multicolumn{1}{c}{[\textit{0.495}]} \\\midrule
    % \multicolumn{1}{c}{[\textit{0.476}]} \\\midrule
    
    24 & 
    % 0.824 & [\textit{0.863}] & 0.825 & \multicolumn{1}{c|}{\textbf{0.877}} & 
    0.370 & 0.358 %& 0.487 
    & 0.374 & 0.431 & 0.499 & [\textit{0.533}] & \textbf{0.552} &  \multicolumn{1}{c}{0.502}  \\ \midrule
    % \multicolumn{1}{c}{0.499}  \\ \midrule
    
    \textbf{Avg.} & 
    % 0.752 & 0.786 & 0.784 & \multicolumn{1}{c|}{\textbf{0.823}} &
    0.532 & 0.559 %& 0.561 
    & 0.564 & 0.574 & 0.624 & 0.629 & [\textit{0.632}] & \multicolumn{1}{c}{\textbf{0.645}}  \\ \bottomrule
    % \multicolumn{1}{c}{\textbf{0.661}}  \\ \bottomrule
    \end{tabular}%
    }
    %  \vspace{-4mm}

\end{table}

\section{Analysis and Discussion}
\label{sec:discussion}
\subsection{Lifecycle-Awareness}
% \vspace*{-1mm}
% In this paper, we present the \acf{AULA-Caps} for spatio-temporal analysis of facial actions. 
The capsule-based computations of the multi-stream \acf{AULA-Caps} allow it to weight the contribution of spatial and spatio-temporal feature capsules towards predicting \ac{AU} activations. If spatial features are more important, for example, in the case of \textit{apex} frames where an \ac{AU} is activated with highest intensity, the model may choose to give precedence to the spatial capsules. For the off-peak intensity frames, for example, originating from the \textit{onset} or \textit{offset} temporal segments where the activation is low, the model may focus more on temporal differences in contiguous frames, captured using the spatio-temporal feature capsules. The ablation study results (see Table~\ref{tab:performance-metrics-ablation-channels}) highlight the individual contribution of spatial (2D) and spatio-temporal (3D) streams where a combination of both, that is, when the model learns to balance these two feature-sets, results in the best model performance. This is consistent with other findings in literature where a combination of spatial and spatio-temporal features results in high performance for \ac{AU} detection~\cite{shao2020spatiotemporal, Yang2019FACS3D}. Interestingly, the windowed computation of spatio-temporal features in \ac{AULA-Caps} performs worse than spatial features, unlike other approaches~\cite{Yang2019FACS3D} where 3D features perform better. This may be due to the choice of a smaller input window ($5$ frames in \ac{AULA-Caps}) unlike~\cite{Yang2019FACS3D} where an entire video is considered for computing spatio-temporal features (see Section~\ref{para:window} for a discussion). 

%  This ability to focus on spatial or spatio-temporal features for any given input is done to make the model \textit{lifecycle-aware}.

% \begin{figure}
%     \centering
%     \includegraphics[width=0.5\textwidth]{Figures/temporal.pdf}
%     \caption{Temporal evaluation of model predictions.}
%     \label{fig:prediction}
%     \vspace{-3mm}

% \end{figure}

% \subsection{Performance}
% \label{sec:performance-discussion}

% Additionally, the co-activation patterns in the data distribution of multiple \acsp{AU} also results in the model not sufficiently learning to predict these \acsp{AU}, confusing them with more frequen other labels. 

% For the BP4D dataset, however, despite the high co-occurrence of different \acsp{AU} (see Figure~\ref{label-BP4D}), the model is sufficiently robust to the imbalanced data distribution due to the overall high number of samples even for under-represented subjects and \acsp{AU}. Furthermore, BP4D dataset provides cleaner and occlusion-free frames where the subjects are recorded mostly in face-centred videos.

\begin{figure}[t]
\centering

\begin{subfigure}{0.235\textwidth}
  \includegraphics[width=\textwidth]{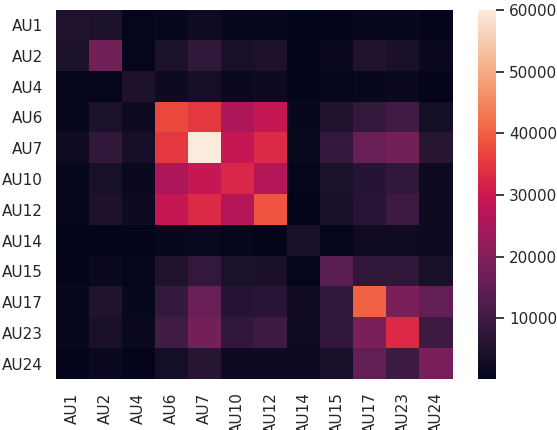}
\caption{GFT} \label{label-GFT}

\end{subfigure}\hfil % <-- added
\begin{subfigure}{0.235\textwidth}
  \includegraphics[width=\textwidth]{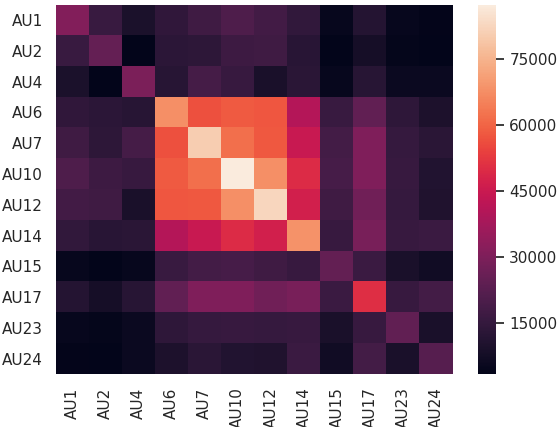}
  \caption{BP4D}
  \label{label-BP4D}
%   \vspace{1mm}

\end{subfigure}
\vspace{-2mm}
\caption{\ac{AU} co-activation heatmaps based on True Labels.}
\label{fig:label-occurrence}
% \vspace*{-3mm}

\end{figure}

\subsection{\ac{AU} Prediction}
\label{sec:performance-discussion}
The \ac{AULA-Caps} model achieves state-of-the-art results for both BP4D and GFT datasets achieving the best overall results on BP4D (see Table~\ref{tab:performance-metrics-BP4D}) while second-best for GFT (see Table~\ref{tab:performance-metrics-GFT}). Despite the good overall performance, individual F1-scores for \acsp{AU}~$1, 4$ and $14$ are quite poor for GFT evaluations. Investigating the data distribution for GFT by plotting the \ac{AU} co-activation heatmap (see Figure~\ref{label-GFT}), we find that certain \acsp{AU} dominate the data distribution. In particular, we see that \acsp{AU} $6, 7, 10$ and $12$ have the highest number of data samples while \acsp{AU} $1, 4$ and $14$, the lowest. In such an imbalanced data distribution, where \acsp{AU} $1, 4$ and $14$ correspond to $<2\%$ of the total samples, the model is unable to learn relevant features that can be used for detecting these \acsp{AU}. The imbalance in data correlates with the performance of the model on individual \ac{AU} (see Table~\ref{tab:performance-metrics-GFT}). A similar imbalance in the data distribution is also witnessed for the BP4D dataset (see Figure~\ref{label-BP4D}) yet, owing to the overall larger number of samples per \ac{AU} label helps mitigate some of these effects for BP4D (see Table~\ref{tab:performance-metrics-GFT}). Furthermore, for the GFT dataset, subjects are recorded interacting with each other in group settings while performing a drink-tasting task which results in a lot of the recorded frames ($\approx23\%$ of the entire dataset) being dropped and not annotated due to occlusions and varying perspectives, impacting the overall data quality as well as distribution. This also negatively impacts the overall results on the GFT database, across the state-of-the-art compared in Table~\ref{tab:performance-metrics-GFT}. BP4D, on the other hand, provides cleaner and occlusion-free frames where the subjects are recorded mostly in face-centred videos resulting in higher performance scores across all the models compared in Table~\ref{tab:performance-metrics-BP4D}. \ac{AULA-Caps} is able to achieve competitive scores on the GFT dataset despite its more complex and challenging settings while outperforming state-of-the-art evaluations on the BP4D dataset.
\begin{figure}
    \centering
    
    \begin{subfigure}{0.23\textwidth}
        \includegraphics[width=\textwidth]{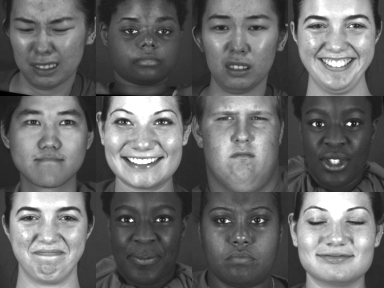}
        \caption{Input \ac{FoI} Images.}
        \label{input-BP4D}
    \end{subfigure}\hfil % <-- added
    % \vspace{1mm}
    \begin{subfigure}{0.23\textwidth}
        \includegraphics[width=\textwidth]{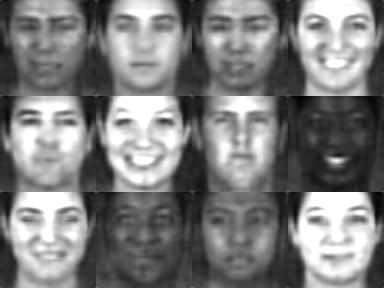}
        \caption{Reconstructed \ac{FoI} Images.} \label{reconstruct-bp4d}
    \end{subfigure}%\hfil % <-- added
    % \vspace*{-1mm}

    \caption{\ac{FoI} Image reconstruction by the Decoder.}
    \label{fig-reconstruct}
% \vspace*{-3mm}

\end{figure}
\begin{figure*}[t]
\centering
%\hfil % <-- added
% \hspace*{4mm}\begin{subfigure}{0.8\textwidth}
%   \includegraphics[width=\textwidth]{Figures/temp_recon.png}
% \end{subfigure}\\
\hspace*{2mm}\begin{subfigure}{0.98\textwidth}
  \includegraphics[width=\textwidth]{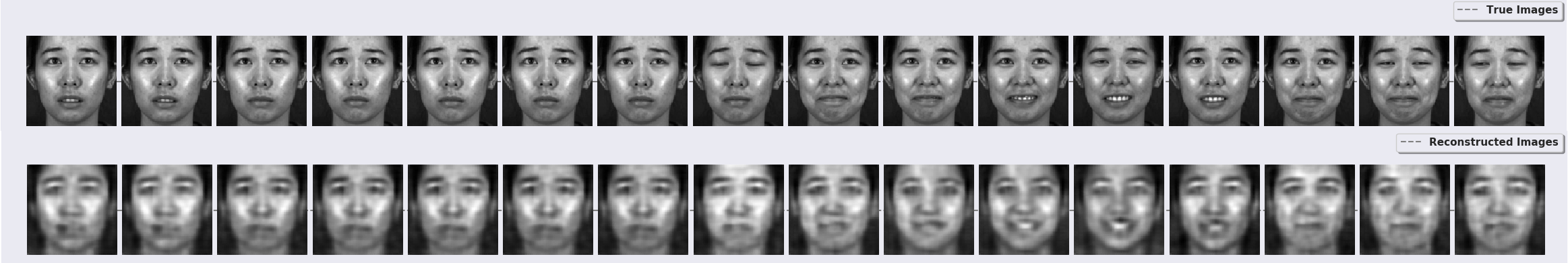}
\end{subfigure}\\
\begin{subfigure}{0.50\textwidth}
  \includegraphics[width=\textwidth]{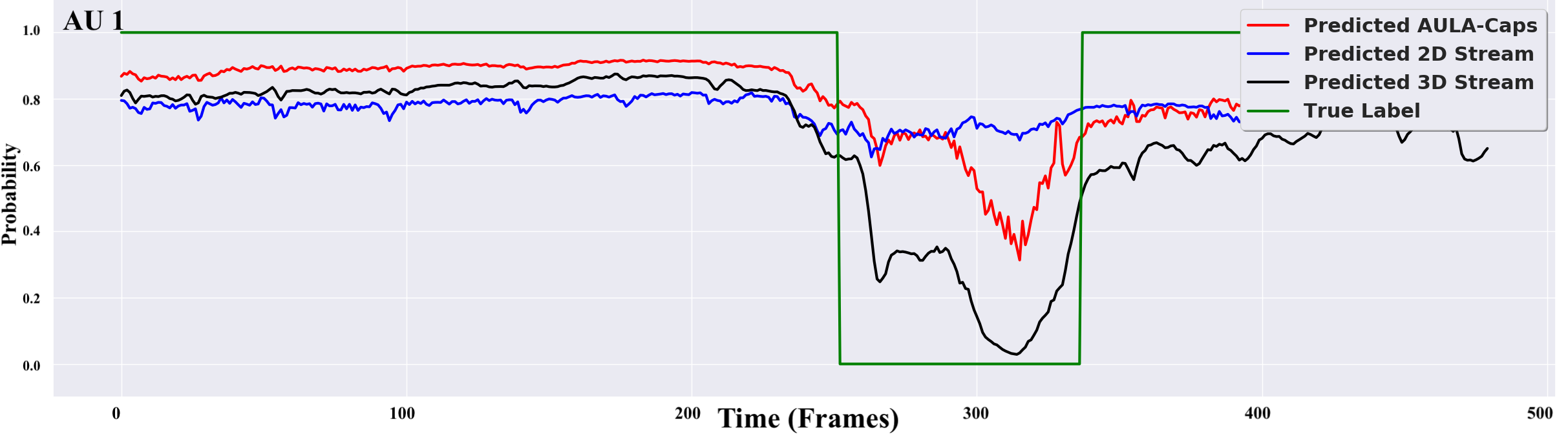}
\end{subfigure}\hfil % <-- added
\begin{subfigure}{0.50\textwidth}
  \includegraphics[width=\textwidth]{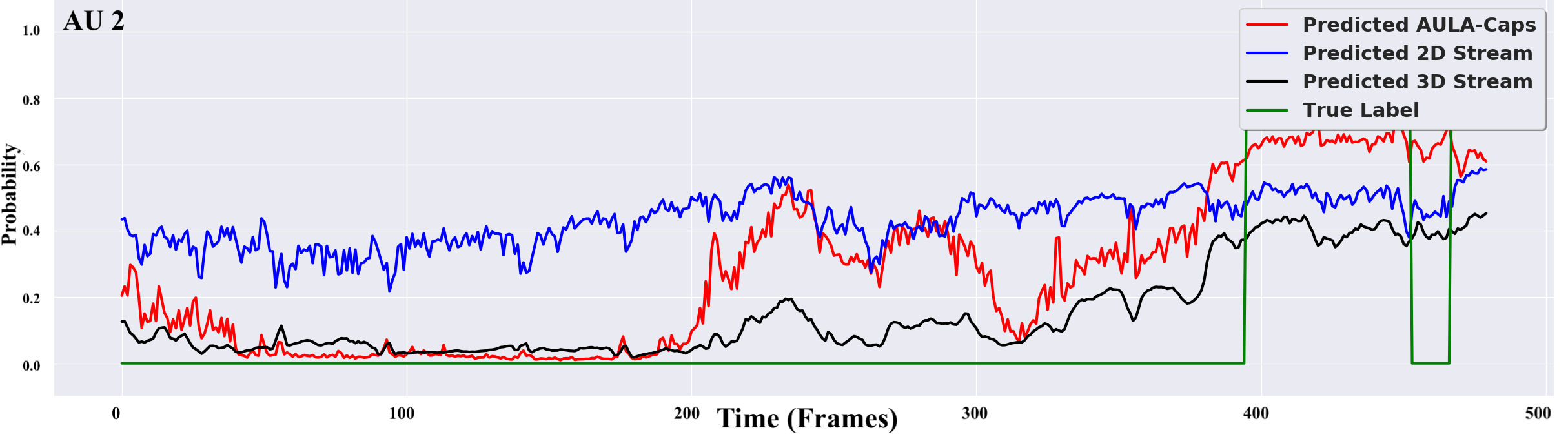}
\end{subfigure}\\%\hfil % <-- added
\begin{subfigure}{0.50\textwidth}
  \includegraphics[width=\textwidth]{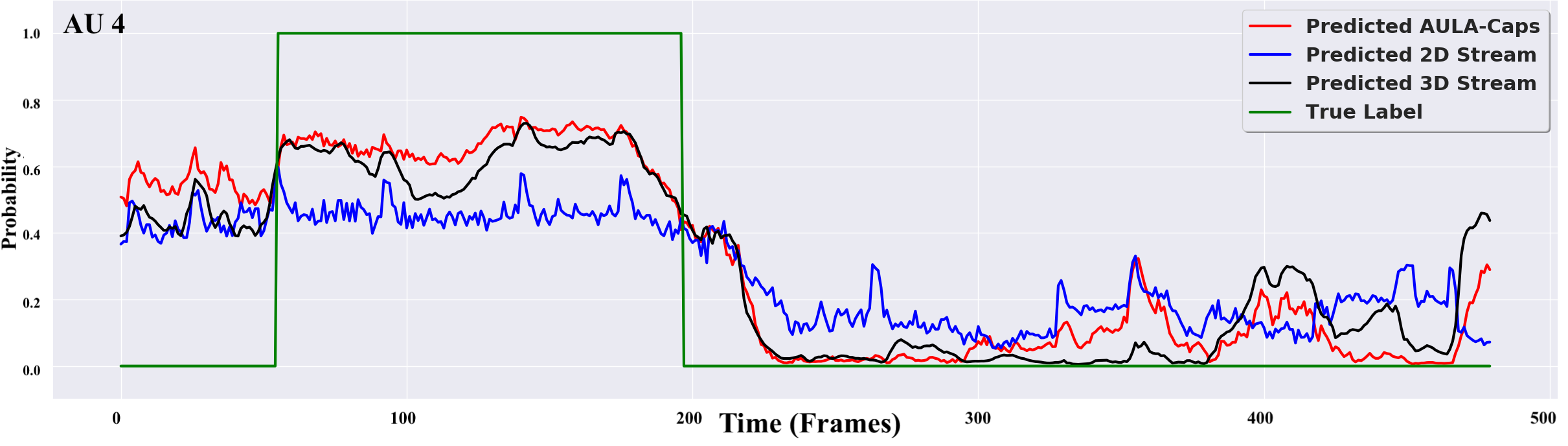}
\end{subfigure}%\hfil % <-- added
\begin{subfigure}{0.50\textwidth}
  \includegraphics[width=\textwidth]{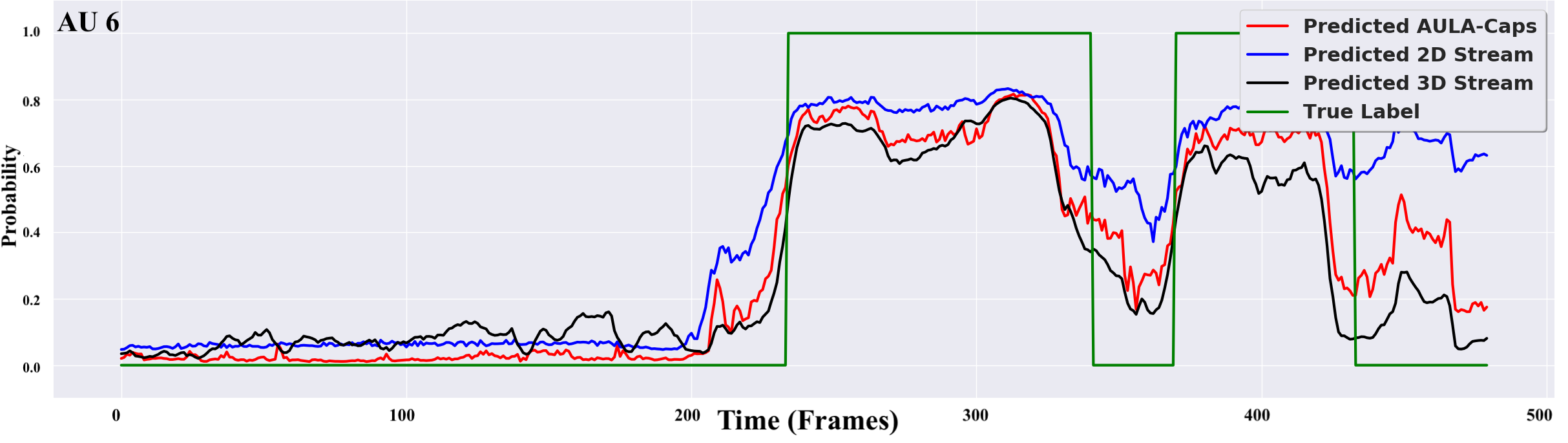}
\end{subfigure}\\%\hfil % <-- added
\begin{subfigure}{0.50\textwidth}
  \includegraphics[width=\textwidth]{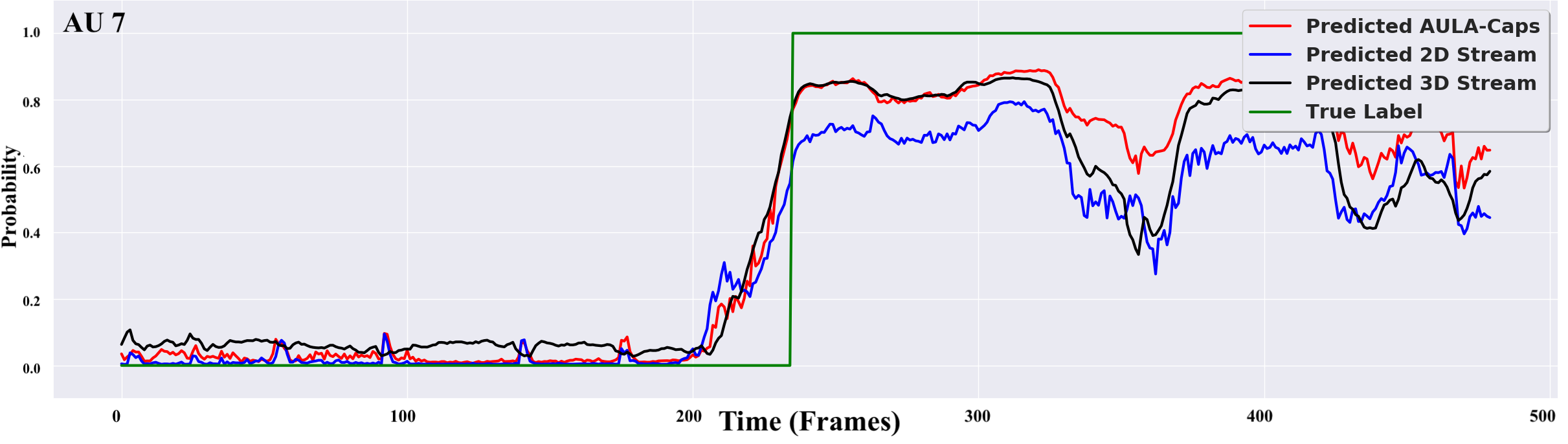}
\end{subfigure}%\hfil % <-- added
\begin{subfigure}{0.50\textwidth}
  \includegraphics[width=\textwidth]{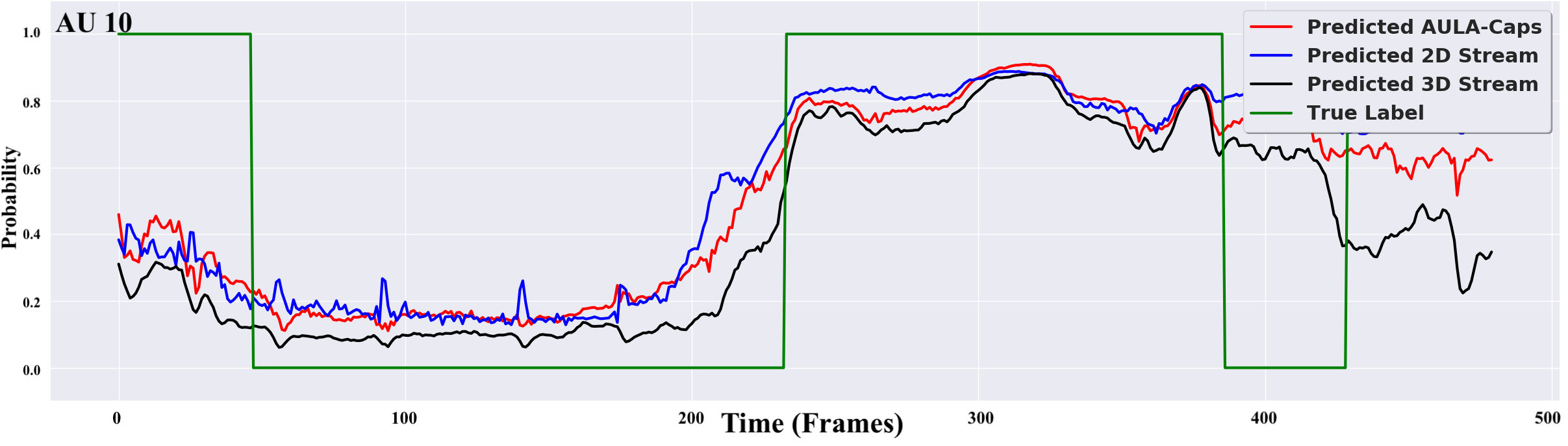}
\end{subfigure}\\
\begin{subfigure}{0.50\textwidth}
  \includegraphics[width=\textwidth]{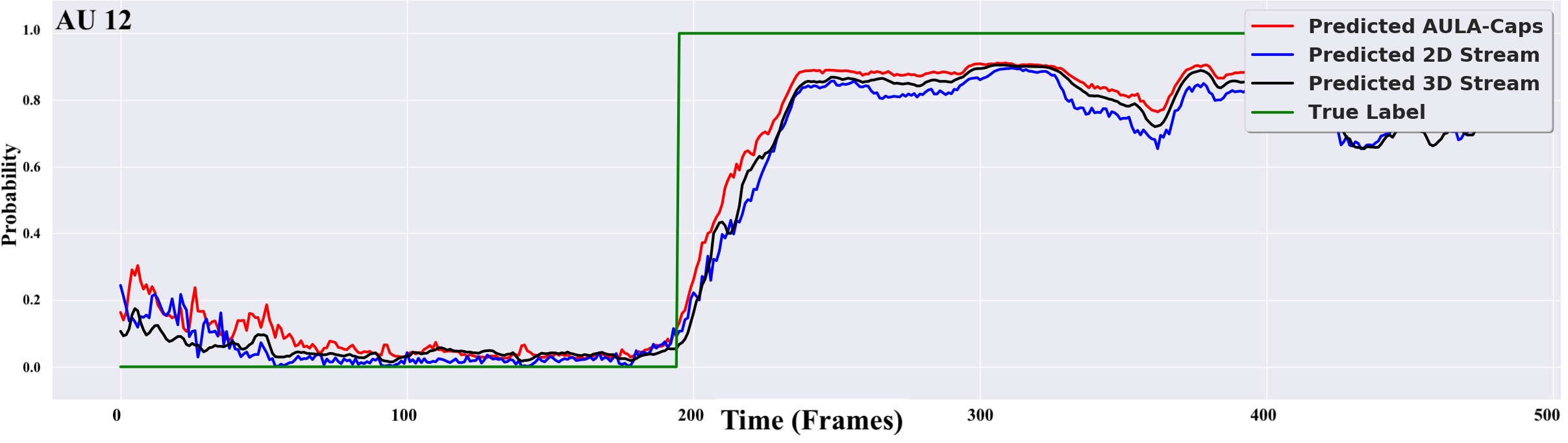}
\end{subfigure}%\hfil % <-- added
\begin{subfigure}{0.50\textwidth}
  \includegraphics[width=\textwidth]{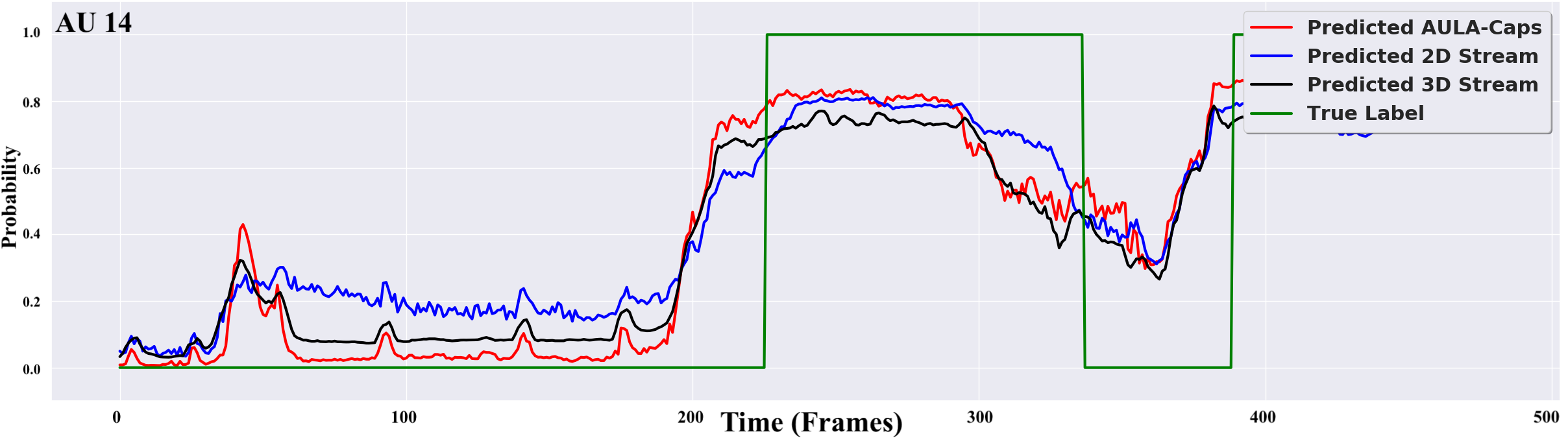}
\end{subfigure}\\
\begin{subfigure}{0.50\textwidth}
  \includegraphics[width=\textwidth]{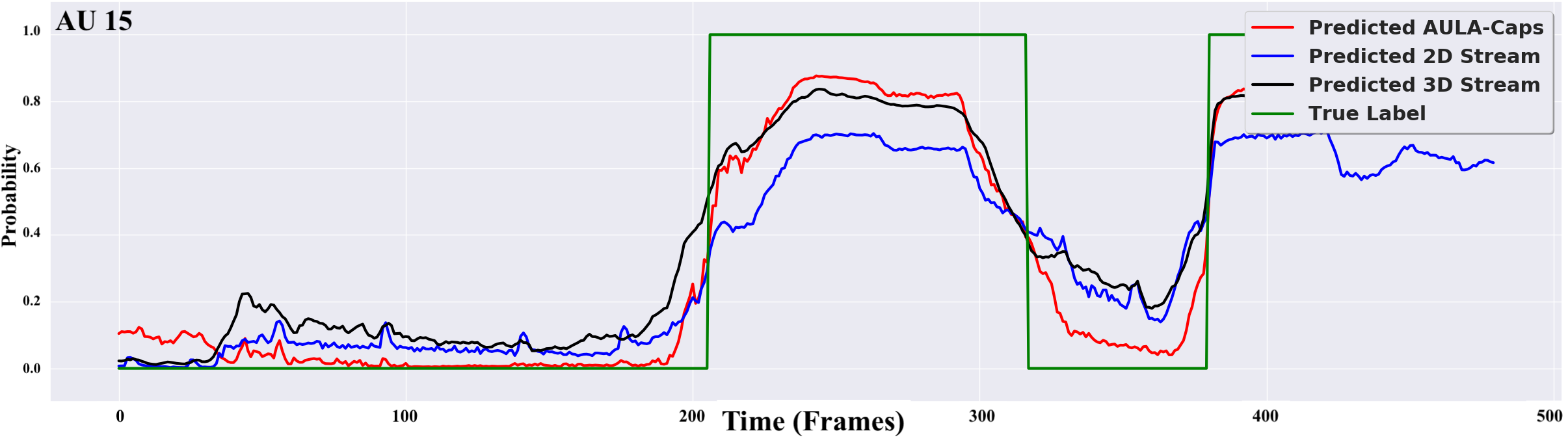}
\end{subfigure}%\hfil % <-- added
\begin{subfigure}{0.50\textwidth}
  \includegraphics[width=\textwidth]{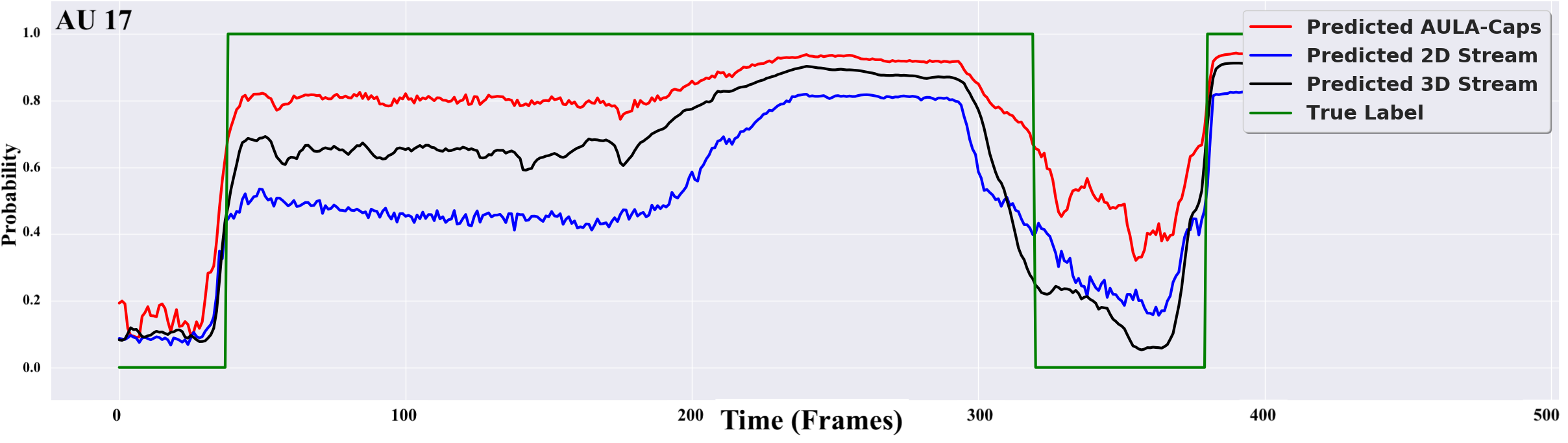}
\end{subfigure}\\
\begin{subfigure}{0.50\textwidth}
  \includegraphics[width=\textwidth]{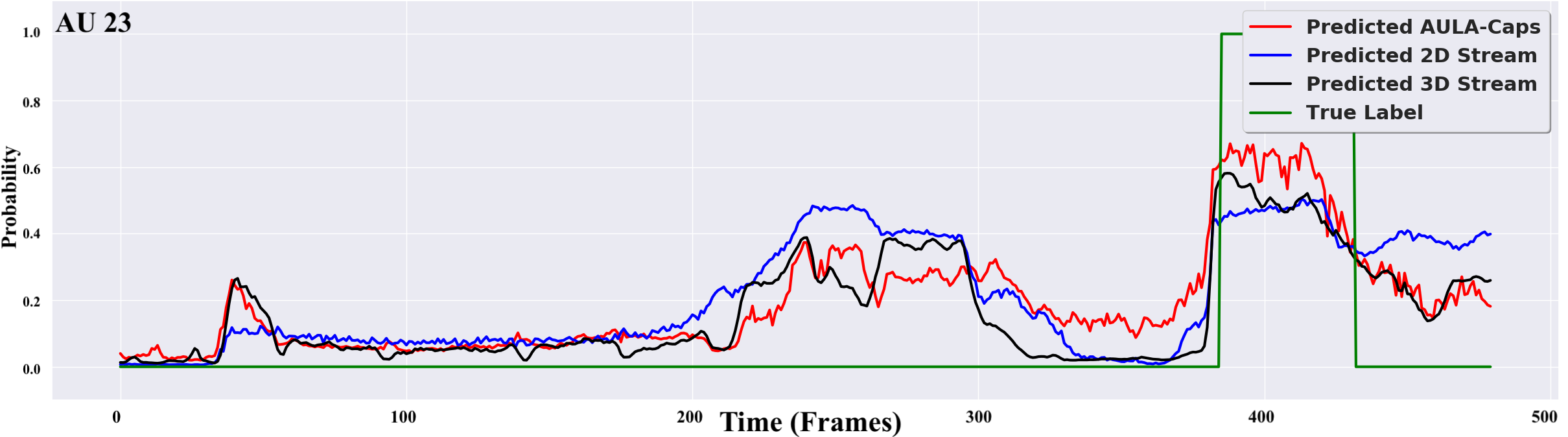}
\end{subfigure}%\hfil % <-- added
\begin{subfigure}{0.50\textwidth}
  \includegraphics[width=\textwidth]{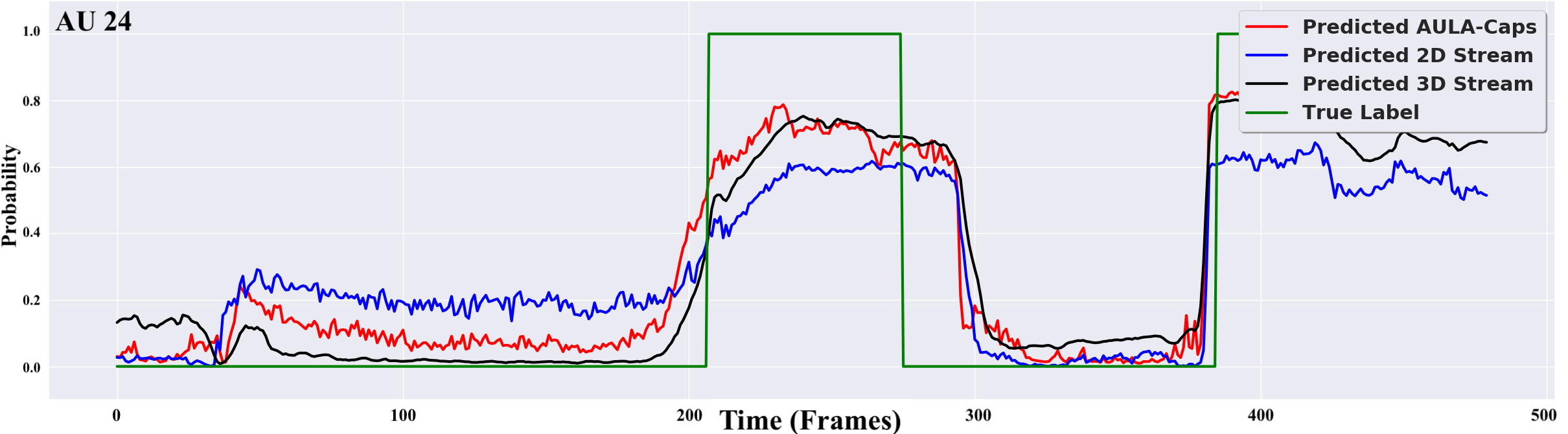}
\end{subfigure}
\caption{Comparing predictions for the $2$D stream, $3$D stream and \ac{AULA-Caps} for the $12$ \acsp{AU} for a sample BP4D video.}
\label{fig:prediction}
%   %\vspace{1mm}

\vspace{-1.5mm}
\end{figure*}

\subsection{Temporal Evaluation} 
\ac{AU} detection evaluations commonly use frame-wise performance metrics for evaluating model performance. However, for automatic \ac{AU} detection, it is also important to evaluate model's performance across time. Considering the data settings in our set-up where video recordings of subjects are examined, predicting \ac{AU} labels in contiguous frames can provide for a \textit{continuous} evaluation of the model. In Figure~\ref{fig:prediction}, we plot, across time, the true labels as well as model predictions, that is, the length of \ac{AU} class capsules for the corresponding \acsp{FoI} depicting the activation probabilities for respective \acsp{AU} for the $2$D stream, $3$D stream and the \ac{AULA-Caps} model. We see that \ac{AULA-Caps} predictions are able to model how the the ground-truth (\ac{AU} activation) varies across time for an entire video. For example, for \ac{AU} $4$, we see the ground truth \ac{AU} activation switching from \textit{absent} to \textit{activated} and then back to \textit{absent} representing its entire lifecycle, while for \acsp{AU} $6, 10, 14, 15, 17$ and $24$ we see this switch occurring multiple times (multiple cycles) within the video. \ac{AULA-Caps} is able to model this switch effectively, predicting \ac{AU} activations efficiently. 

Furthermore, Figure~\ref{fig:prediction} also shows the predictions from the $2$D and $3$D streams where it can be seen that the $3$D stream, on average, is able to better model the changing dynamics of \ac{AU} activations as compared to the $2$D streams, especially in regions where the ground truth switches from absent to activated or vice-versa. Yet, the $2$D stream outperforms the average frame-based performance of the $3$D stream across all the videos and all \acsp{AU}. As frame-based performance evaluations for \ac{AU} detection algorithms only report the \textit{average} frame-based F1-scores, they ignore temporal correspondences, with respect to model performance, which are commonly examined for continuous affect perception such as arousal-valence prediction~\cite{Nicolaou2011Continuous,Yannakakis2017Ordinal}. Yet, these can be really beneficial for understanding real-time model performance, underlining its applicability for \textit{automatic} \ac{AU} prediction in real-world applications.

\subsection{Visualisations}
% \vspace*{-2mm}
\paragraph{Image Reconstruction:}
The decoder is used to regularise learning by ensuring the model learns task-relevant features. Additionally, the reconstructed images enable a visual interpretation of the learnt features. In \ac{AULA-Caps}, we propose a convolution-based decoder model as opposed to the densely-connected decoder originally proposed by Sabour \etal~\cite{CAPS2017}. The decoder is able to generate reconstructed images using a much `lighter' network ($\approx2$M parameters \vs $\approx 10$M~\cite{CAPS2017}) without compromising on the quality of the images generated, as can be seen in Figure~\ref{fig-reconstruct}. 

The data imbalance problem is witnessed in the reconstructed images as well where \acsp{FoI} for certain under-represented subjects and labels are reconstructed incorrectly. For example, for the images at (row $1$, column $2$) and (row $2$, column $1$) in both Figures~\ref{input-BP4D} and~\ref{reconstruct-bp4d}, the model reconstructs \textit{generic} \textit{mean} faces representing the corresponding \ac{AU} labels, with a visible bias in terms of ethnicity and/or gender.

\begin{figure}[t]
    \centering
    % \begin{subfigure}{0.48\textwidth}
    %     \includegraphics[width=\textwidth]{Figures/heatmap_layer_primarycap_conv2d_saliency_GFT.png}
    %     \caption{Visualising Saliency Maps on a GFT Sample.}
    %     \label{saliency-GFT}
    % \end{subfigure} \hfil
    % <-- added
    % \begin{subfigure}{0.48\textwidth}
        \includegraphics[width=0.5\textwidth]{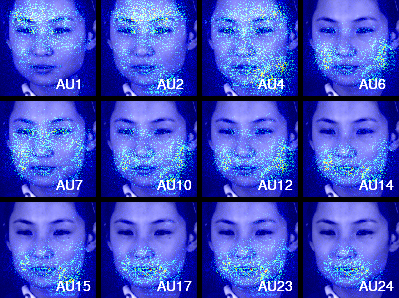}
        % \caption{Visualising Saliency Maps on a BP4D Sample.} \label{saliency-BP4D}
    % \end{subfigure} % <-- added
    % \vspace{-2mm}

    \caption{Saliency Maps generated using \textit{guided backpropagation} of gradients corresponding to each \ac{AU} label.
    }
    \label{fig:saliency-maps}
    \vspace{-2mm}
\end{figure}

% \vspace*{-3mm}
\paragraph{Visualising Saliency Maps:}
% \label{sec:discussion}
Visualising learnt features enables us to understand what the model pays attention to while making its predictions. In Figure~\ref{fig:saliency-maps}, we see the Saliency Maps~\cite{Simonyan14a} generated by visualising the pixels in the \acsp{FoI} that contribute most to model predictions. As desired, for different \acsp{AU} the model learns to focus on different regions of the face, in line with Table~\ref{tab:AU_labels}. For example, for \acsp{AU} $1$ and $2$ it focuses more on the \textit{forehead} and \textit{eyebrows}, for \acsp{AU} $12$ and $14$ it focuses on \textit{cheeks} while for \acsp{AU} $23$ and $24$, it focuses on the \textit{nose} and \textit{mouth}. For certain \acsp{AU} however, we see additional activity in other `non-relevant' regions of the face as well. For example, for \ac{AU} $4$, we see activity in the lower region of the face near the mouth and cheeks. This can be due to the co-occurrence pattern observed in the label distribution where samples containing \ac{AU} $4$ also encode activity for \ac{AU} $7$ and \ac{AU} $17$, as illustrated in Figure~\ref{fig:label-occurrence}. Understanding and evaluating such co-occurrence patterns can be important to improve model predictions for \ac{AU} activations~\cite{li2019semantic,Tong2008BN} as knowing if a certain \ac{AU} is activated or not can help improve model predictions for other correlated \acsp{AU}.

\section{Conclusion}

Our experiments with the \ac{AULA-Caps} highlight the importance of combining spatial and spatio-temporal feature learning for automatic \ac{AU} prediction. Evaluating the temporal evolution of \ac{AU} activation positively impacts model performance and allows for the dynamic evaluation of \ac{AU} activity in a continuous manner. This is inline with other findings~\cite{Pantic2005Detecting, shao2020spatiotemporal}. Furthermore, capsule-based computations in the spatial stream enable learning local spatial relationships within an image frame corresponding to the different face regions while in the spatio-temporal stream they are able to learn temporal dependencies based on how these spatial relationships evolve across time. Combining such features allows the model to learn where to focus in an image while also being sensitive to the \ac{AU} activation lifecycle. %, learning whether to focus more on spatial or spatio-temporal features.

% \vspace*{-1mm}
\subsection{Limitations and Future Work}
% Despite the model achieving state-of-the-art performance for several \ac{AU} labels, here we highlight certain key limitations that need to be addressed.

% \vspace*{-1mm}
\paragraph{Choosing the Right Window for Context:}
\label{para:window}
As the model evaluates \ac{AU} activity across a window of input frames, it is highly sensitive to how these windows are processed. For GFT, we see that due to occlusions and complex recording conditions, several frames are dropped randomly as no \ac{AU} activity is annotated for those frames. This impacts model performance resulting in poor performance for certain \acsp{AU} such as \ac{AU} $1, 4$ and $14$. Additionally, the size of the input window may impact model performance differently for the different \acsp{AU}. For some \acsp{AU} the lifecycle is much longer than the others, for example, \ac{AU} $12$ (smile) \vs \ac{AU} $45$ (blink), and thus wider windows are expected to improve performance. In our experiments, however, we optimised the window-size based on the highest overall F1-score. Further experimentation is needed to investigate which window-sizes work best for different \acsp{AU}. Also, learning to dynamically adapt such windows based on \ac{AU} activity may offer improvements. Lu \etal~\cite{Lu2020BMVC} provide an insightful approach to address such problems by focusing on the temporal consistency in video sequences rather than relying on pre-defined window-sizes. They randomly assign \textit{anchor frames} in input sequences and apply self-supervised learning to encode the temporal consistency of an input video sequence compared to this anchor frame. This robustly captures temporal dependencies in facial activites, improving \ac{AU} detection performance.

\vspace*{-3mm}
\paragraph{Imbalanced Data Distributions:}
Another problem faced by most approaches is the imbalanced label distribution of the datasets. In Figure~\ref{fig:label-occurrence}, we see that \acsp{AU} $6, 7, 10$ and $12$ dominate the data distributions, resulting in the models performing worse on scarce labels such as \acsp{AU} $1, 4$ and $14$. It is important to address this imbalance either at the data-level by recording evenly distributed datasets that offer a fairer comparison of models or by including mitigation strategies that handle biases arising from such imbalances~\cite{CHARTE2015MLSMOTE, Oksuz2020PAMI, Xu2020InvestigatingBA}. Furthermore, understanding \ac{AU} co-activations can provide additional contextual information to improve performance on scarce \ac{AU} samples~\cite{li2019semantic,Tong2008BN,Zhao2016Deep}.

\section*{Acknowledgement}

{\noindent}N.~Churamani is funded by the EPSRC grant EP/R$513180$/$1$ (ref.~$2107412$). H.~Gunes is funded by the European Union's Horizon $2020$ research and innovation programme, under grant agreement No. $826232$. S.~Kalkan is supported by Scientific and Technological Research Council of Turkey (T\"UB\.ITAK) through BIDEB $2219$ International Postdoctoral Research Scholarship Program. The authors also thank Prof Lijun Yin from the Binghamton University (USA) for providing access to the BP4D Dataset; Prof Jeff~Cohn and Dr Jeffrey~Girard from the University of Pittsburgh (USA) for providing access to the GFT dataset.

%H. Gunes is supported by the EPSRC (grant ref.~EP/R030782/1) and the Alan Turing Institute Faculty Fellowship (G102185).
%------------------------------------------------------------------------

\begin{acronym}
\acro{AC}{Affective Computing}
\acro{AI}{Artificial Intelligence}
\acro{AU}{Action Unit}
\acro{AULA-Caps}{Action Unit Lifecycle-Aware Capsule Network}
\acro{BP4D}{Binghamton-Pittsburgh 3D Dynamic (4D) Spontaneous Facial Expression Database}
\acro{CL}{Continual Learning}
\acro{CNN}{Convolutional Neural Network}
\acro{FACS}{Facial Action Coding System}
\acro{FER}{Facial Expression Recognition}
\acro{FoI}{Frame-of-Interest}
\acro{GAN}{Generative Adversarial Network}
\acro{GFT}{Group Formation Task}
\acro{HRI}{Human-Robot Interaction}
\acro{ML}{Machine Learning}
\acro{MLP}{Multilayer Perceptron}
\acro{LSTM}{Long Short-Term Memory}
\acro{ReLU}{Rectified Linear Unit}
\acro{SSL}{Self Supervised Learning}
\end{acronym}
% \newpage
% \balance

{\small
\bibliographystyle{ieee_fullname}
\bibliography{main}
}
\end{document}